\documentclass[sigconf]{acmart}
\usepackage{booktabs} 
\usepackage[utf8]{inputenc}
\usepackage{subcaption}
\usepackage{microtype}
\usepackage{enumitem}

\copyrightyear{2017}
\acmYear{2017}
\setcopyright{acmlicensed}
\acmConference[GECCO '17]{the Genetic and Evolutionary Computation Conference 2017}{July 15--19, 2017}{Berlin, Germany}
\acmPrice{15.00}
\acmDOI{http://dx.doi.org/10.1145/3071178.3071292}
\acmISBN{978-1-4503-4920-8/17/07}

\DeclareMathOperator{\var}{\textrm{var}}

\DeclareMathOperator{\cov}{\textrm{cov}}
\def\mean#1{\left< #1 \right>}

\begin{document}
\title{Neuroevolution on the Edge of Chaos}

\author{Filip Matzner}
\affiliation{%
  \institution{Charles University in Prague\\
               Faculty of Mathematics and Physics}
  \streetaddress{Malostranské náměstí 25}
  \city{Prague} 
  \state{Czech Republic} 
  \postcode{118 00}}
\email{floop@floop.cz}

\begin{abstract}
\textit{Echo state networks} represent a special type of \textit{recurrent neural networks}. Recent papers stated that the echo state networks maximize their computational performance on the transition between \textit{order} and \textit{chaos}, the so-called \textit{edge of chaos}. This work confirms this statement in a comprehensive set of experiments. Furthermore, the echo state networks are compared to networks evolved via \textit{neuroevolution}. The evolved networks outperform the echo state networks, however, the evolution consumes significant computational resources. It is demonstrated that echo state networks with local connections combine the best of both worlds, the simplicity of random echo state networks and the performance of evolved networks. Finally, it is shown that evolution tends to stay close to the ordered side of the edge of chaos.

\end{abstract}

%
%

\ccsdesc[500]{Computing methodologies~Neural networks}
\ccsdesc[500]{Mathema-tics of computing~Time series analysis}
\ccsdesc[100]{Mathema-tics of computing~Information theory}

\keywords{{recurrent neural networks},
{echo state networks},
{neuroevolution},
{edge of chaos},
{phase transition}}

\maketitle

\section{Introduction}

Even though recurrent neural networks seem to provide an efficient way to process data in biological organisms, artificial models of recurrent networks are currently considered to be more difficult to train than their feed-forward counterparts \cite{pascanu2013difficulty}. In feed-forward networks, input data propagate through the network in the feed-forward manner only, hence the data pass through each neuron only once. In the case of recurrent neural networks, the data might flow through the network back and forth for an extended time period, each neuron might process the data multiple times, and the data from different time points can be combined to build various time dependent relations. In other words, recurrent networks implicitly benefit from memory capability, that allows them to solve various non-Markovian tasks\footnote{In Markovian tasks the input in a single time point provides a complete state information necessary to solve the task, opposite to the non-Markovian tasks, where a longer history of input values might be required. For instance, deducing whether a car has reached its known destination point, knowing the position of the car in each time step, is a simple Markovian task. However, if only the velocity of the car in each time step is known, the task is non-Markovian, since the velocity has to be integrated over time to calculate the position.} without being explicitly provided by additional inputs.

Recurrent networks are harder to train, however, they provide a set of very desirable properties. To overcome the cumbersome training, \citeauthor{jaeger2001echo} developed a new approach known as \textit{echo state networks} \cite{jaeger2001echo}. This model significantly speeds up training and avoids most of the training pitfalls at the cost of decreased adaptation ability. The key element in echo state networks is a large, \textit{randomly} generated recurrent network. Echo state networks rely on the assumption that this large random network nonlinearly transforms the input into so many variations that extraction of useful information becomes simple.

\begin{figure}
  \centering
  \begin{subfigure}[t]{0.32\linewidth}
    \includegraphics[width=\textwidth]{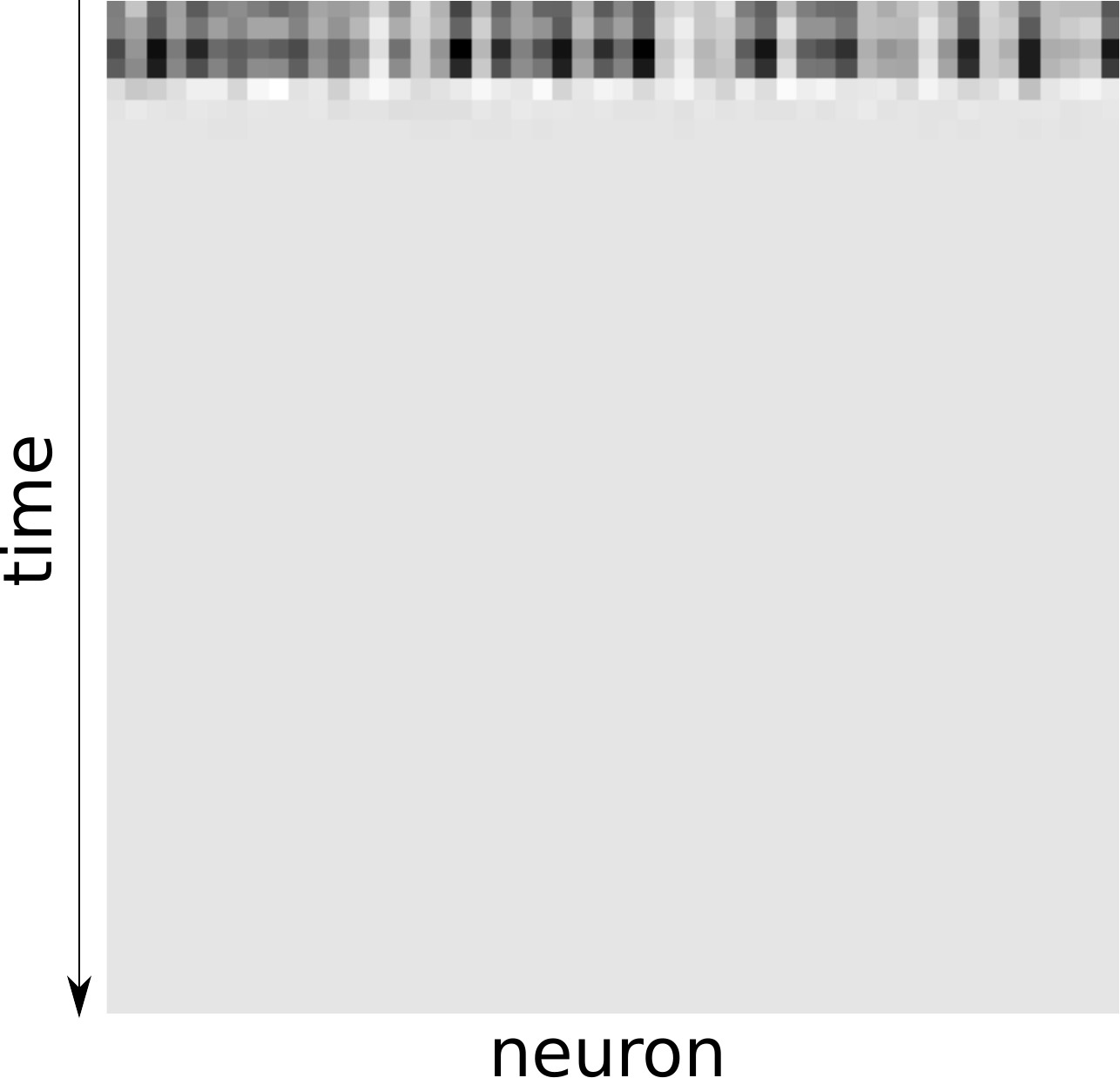}
    \captionsetup{width=0.8\textwidth}
    \caption{Ordered net activity tends to die out.}
    \label{fig:ordered_net}
  \end{subfigure}
  \begin{subfigure}[t]{0.32\linewidth}
    \includegraphics[width=\textwidth]{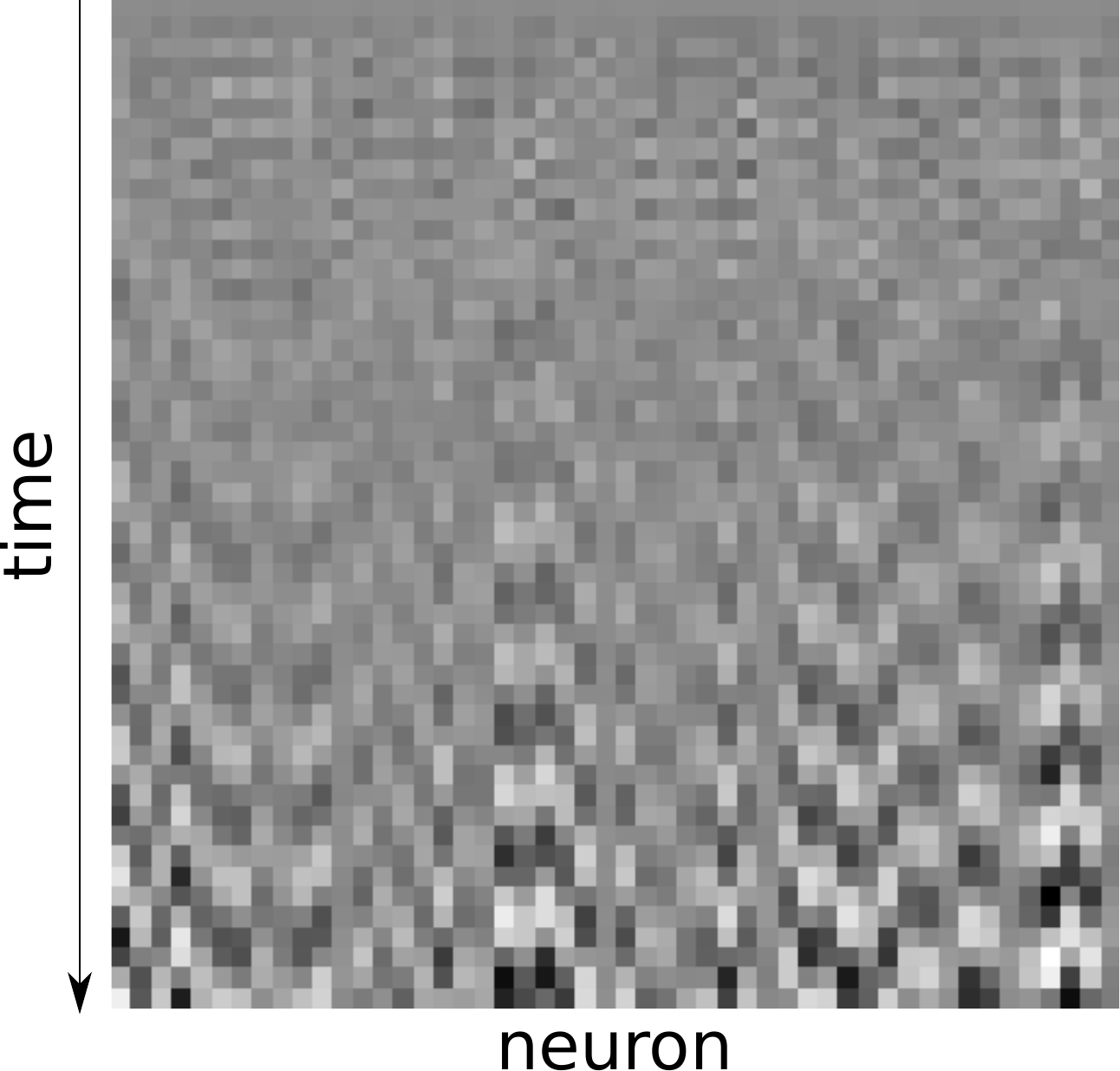}
    \captionsetup{width=0.8\textwidth}
    \caption{Edge of chaos net produces repeating patterns.}
    \label{fig:edge_of_chaos_net}
  \end{subfigure}
  \begin{subfigure}[t]{0.32\linewidth}
    \includegraphics[width=\textwidth]{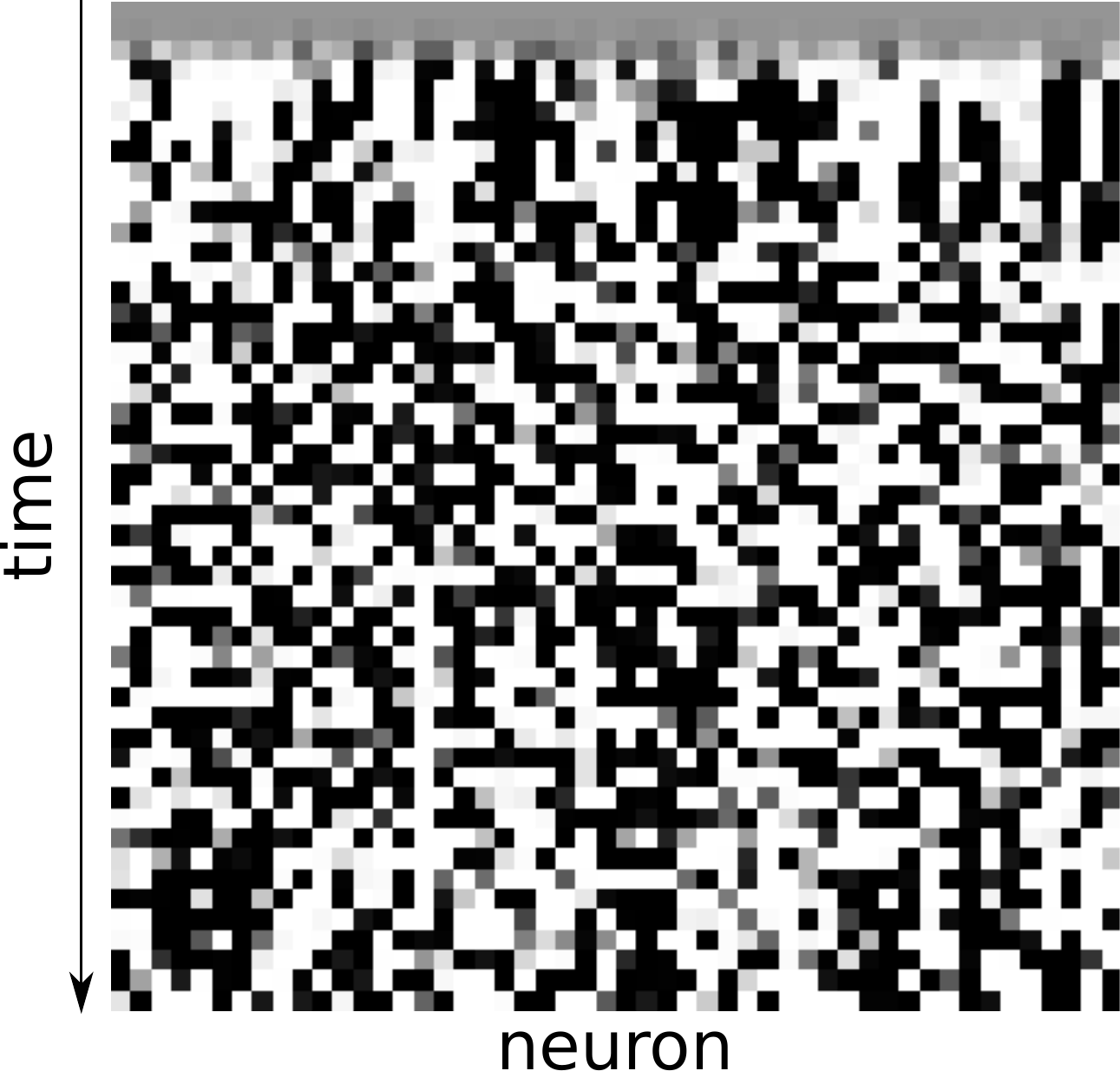}
    \captionsetup{width=0.8\textwidth}
    \caption{Chaotic net output resembles random white noise.}
    \label{fig:chaos_net}
  \end{subfigure}
  \caption{Three echo state networks tuned for ordered, edge of chaos and chaotic dynamics (left to right). Each row represents the activities of 50 neurons in the recurrent part of the network. The network was driven by an input sequence of 5 random numbers and 45 zeros.}
\end{figure}

There are, however, some restrictions to the random weights in order for the recurrent network to behave reasonably. When the weights are too large, the network's output resembles a white noise and it is called to have a \textit{chaotic dynamics} (Figure~\ref{fig:chaos_net}). In the opposite case, where the weights are too small, the activity of the network tends to die out and it is called to have an \textit{ordered dynamics} (Figure~\ref{fig:ordered_net}). In neither of the two cases is it possible to extract anything useful out of the network. According to the papers by \citeauthor{bertschinger2004real} \cite{bertschinger2004real} and \citeauthor{boedecker2012information} \cite{boedecker2012information}, the best properties are provided when the recurrent network dynamics is on the very transition between order and chaos (Figure~\ref{fig:edge_of_chaos_net}). This particular transition was given the name \textit{edge of chaos} \cite{langton1990edgeofchaos}.

Similar phenomena has been observed also in the fields of cellular automata \cite{langton1990edgeofchaos}, boolean networks \cite{kauffman1993origins}, spiking networks \cite{legenstein2007edgeofchaos}, and even biological cortical circuits \cite{beggs2008criticality}.

\section{Problem Specification}

Our primary goal is to evolve the recurrent part of echo state networks using neuroevolution and analyze the computational power of such evolved networks and their relation to the edge of chaos. Both subjects, the computational power and the edge of chaos relation, will be compared with the corresponding properties of the original, pure random, echo state networks.

Echo state networks in combination with the edge of chaos suggest an idea of how biological networks might achieve such an unbeaten performance. Instead of training each neural synapse separately, the brain tissue might grow more or less randomly and still obtain great results by remaining on the edge of chaos. Unfortunately, the papers by \citeauthor{bertschinger2004real} \cite{bertschinger2004real} and \citeauthor{boedecker2012information} \cite{boedecker2012information}, which evaluate the performance of echo state networks on the edge of chaos, only consider pure random networks where all pairs of neurons have the same probability of being connected. Such networks consist of a structure with no regularities, no repeating patterns and no locality dependencies. Such a model does not comply with the knowledge of biological neural tissue, that might have a higher degree of regularity \cite{squire2013neuroscience}.

To allow evolution of biologically plausible networks, we will use the HyperNEAT algorithm \cite{gauci2007hyperneat}, which provides the means to build complex regular structures. We are interested whether the biologically plausible networks will perform comparably to their pure random counterparts and whether their performance will relate to the edge of chaos.

Before the main experiment, we will replicate the original results of \citeauthor{bertschinger2004real} \cite{bertschinger2004real} and \citeauthor{boedecker2012information} \cite{boedecker2012information} who propose that computational power of echo state networks is maximized on the edge of chaos.

\section{Methods}
\label{sec:used_methods}

In this section, all the methods used in the experiments will be explained to a greater depth.

\subsection{Echo State Networks}

\begin{figure}
	\centering
	\includegraphics[width=\linewidth]{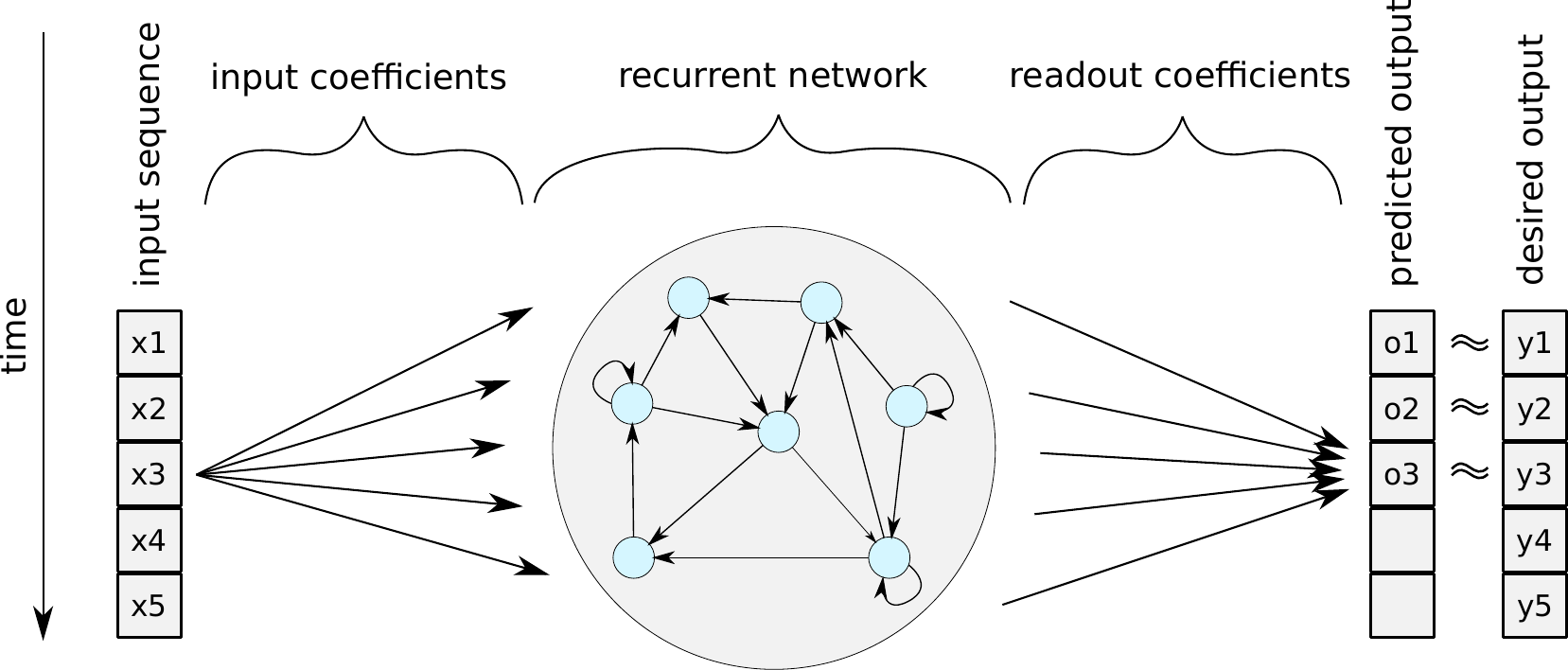}
  \caption{Illustration of an echo state network.}
  \label{fig:esn_intro}
\end{figure}

An echo state network, defined by \citeauthor{jaeger2001echo} \cite{jaeger2001echo}, consists of a recurrent network with weight matrix $W$, a vector of \textit{input coefficients} $\overrightarrow{w}_{\textrm{in}}$, and a vector of \textit{readout coefficients} $\overrightarrow{w}_{\textrm{out}}$ (Figure~\ref{fig:esn_intro}). The activations of the neurons in the recurrent network, the input value, and the output value in time $t$ are denoted by $\overrightarrow{a}^t$, $x^t$ and $o^t$, respectively. The activations and the output value are calculated as follows:
\begin{align*}
  \textstyle
  \overrightarrow{a}^t &= \textrm{tanh}(W \overrightarrow{a}^{t-1} + \overrightarrow{w}_{\textrm{in}} x^t) \,, \\
  o^t &= \overrightarrow{w}_{\textrm{out}} \cdot \overrightarrow{a}^t \,.
  \label{eq:esn}
\end{align*}

The recurrent network and the input coefficients are generated randomly and never change, the only part that is trained for the given problem are the readout coefficients. They are chosen so that the predicted sequence and the desired output sequence minimize their squared distance.

For the echo state network to work properly, the recurrent network cannot be completely random. Instead, it shall have a so-called \textit{echo state property} (sometimes called \textit{fading memory}). Informally, it means that the state of the netwok only depends on a finite history of its inputs. A more formal definition will not be provided, as it is equivalent to the definition of a network with ordered dynamics described in the next section \cite{bertschinger2004real}. More information about echo state networks and their training can be found in the original paper by \citeauthor{jaeger2001echo} \cite{jaeger2001echo}.

\subsection{Chaotic and Ordered Dynamics}
\label{sec:overview_chaotic_ordered_dynamics}

Let us briefly introduce \textit{chaotic} and \textit{ordered} dynamics. A system in which a sufficiently small \textit{perturbation of initial parameters} disappears in a \textit{finite time} is called to be ordered. A system in which a perturbation amplifies is called to be chaotic.

In this section, we will describe a measure of chaoticity called \textit{Lyapunov exponent} (denoted by $\lambda$) in the context of neural networks. Its rationale is to let a neural network run from two slightly perturbed initial states and measure the distance between the two network states from that moment on. If the two network states tend to converge, the system is in the ordered phase and $\lambda < 0$. If they diverge, the system is in the chaotic phase and $\lambda > 0$. The \textit{edge of chaos} lies right in the middle, where the two states tend to keep the same distance from each other and $\lambda \approx 0$.

The formal definition of Lyapunov exponent $\lambda$ is following:
\begin{equation*}
  \textstyle
  \lambda = \lim_{t \to \infty} \left( {1}/{t} \right) \ln \left( {\gamma_t}/{\gamma_0} \right) \,,
  \label{eq:lyapunov}
\end{equation*}
where $\gamma_t$ is the distance of the two initially perturbed states in time $t$ and $\gamma_0$ is the distance of the initial states, i.e., the size of the perturbation.

We will adopt the algorithm by \citeauthor{sprott1997lyapunov} \cite{sprott1997lyapunov} for numerical estimation of Lyapunov exponent. Let us explain the process.

\begin{figure}
	\centering
	\includegraphics[width=0.6\linewidth]{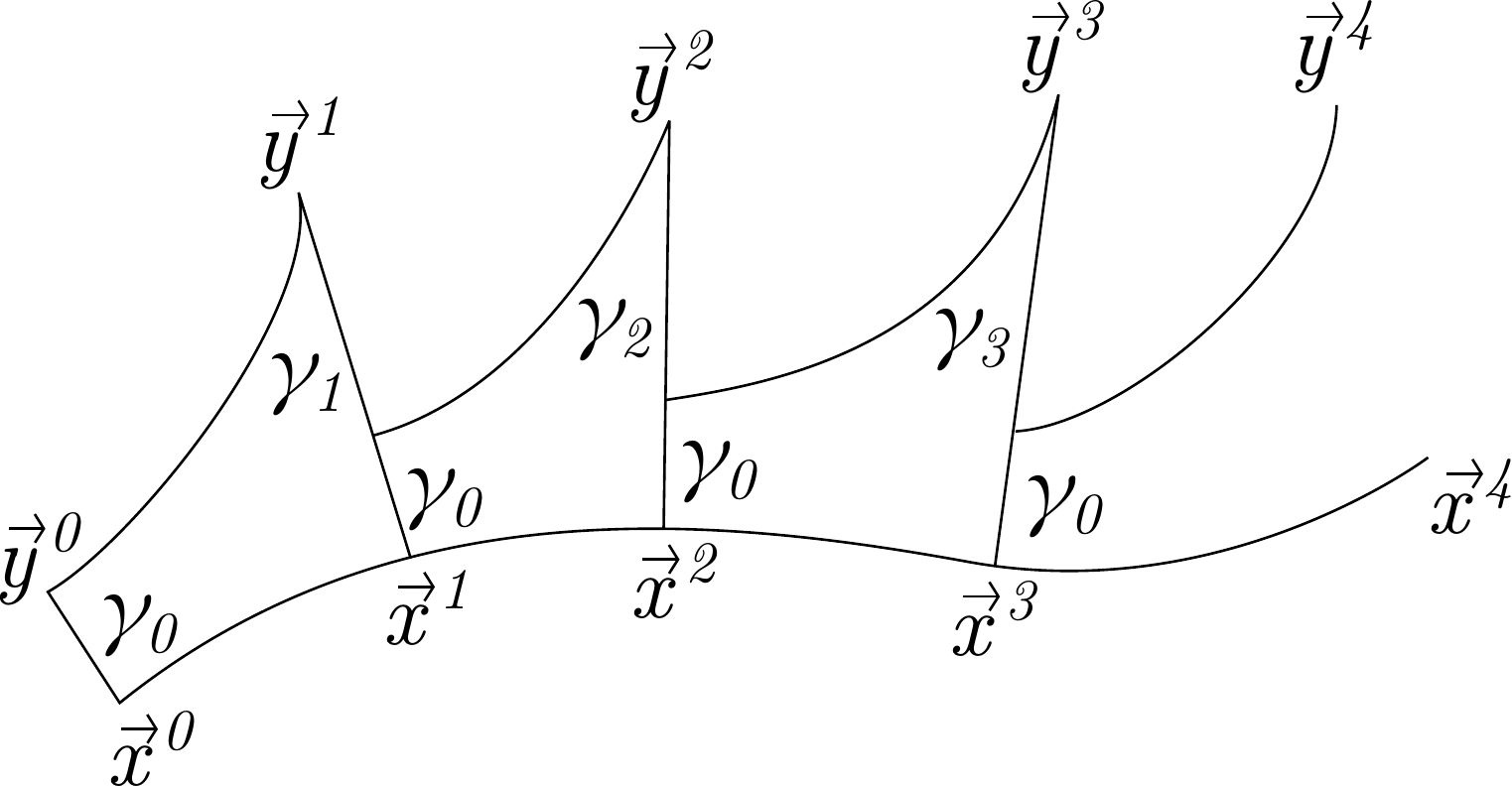}
  \caption{Illustration of the Lyapunov exponent algorithm. After each time step, the distance between the original and the perturbed network is recorded and normalized to the initial value of $\gamma_0$.}
  \label{fig:lyapunov_steps}
\end{figure}

\begin{enumerate}[leftmargin=*]
  \item A random sequence of 2000 values is generated to drive the neural network.
  \item The network is run for 1000 time steps and its outputs are discarded. This action is performed for the network to stabilize.
  \item After the 1000 steps, the network is duplicated. In the second network, the activation value of \textit{one of its neurons} is perturbed by the value of $\gamma_0$ (in this paper $\gamma_0 = 10^{-12}$).
  \item Both the networks proceed one time step forward and the distance between their states is calculated. Formally, the distance will be denoted by $\gamma_t = \left| \left| \overrightarrow{x}^t - \overrightarrow{y}^t \right| \right|$, where $\overrightarrow{x}^t$ and $\overrightarrow{y}^t$ are the activations of the neurons in the first and the second network respectively and $|| \cdot ||$ denotes the Euclidean distance. This distance is recorded for later use.
  \item The state of the perturbed network is normalized to the initial distance of $\gamma_0$, i.e., $\overrightarrow{y}^t = \overrightarrow{x}^t + \left( {\gamma_0}/{\gamma_t} \right) \left( \overrightarrow{y}^t - \overrightarrow{x}^t \right)$. This action is performed because the activation value of a neuron usually has a limited range (e.g., $[0; 1]$ for sigmoidal transfer function) and therefore, the distance between the states is also limited. This action ensures that the two states do not diverge close to this limit and also avoids numerical overflows. Figure~\ref{fig:lyapunov_steps} visualizes this operation.
  \item Repeat steps 4 and 5 until the end of the input sequence.
  \item Return to (3) and choose a different neuron to be perturbed.
\end{enumerate}
This process is repeated for each neuron in the network. The final Lyapunov exponent is the average logarithm of the distance of the two trajectories, averaged over all neurons:
\begin{equation*}
  \textstyle
  \lambda = \mean{\ln \left( {\gamma^n_t}/{\gamma_0} \right) }_{t, n} \,,
  \label{eq:lyapunov_final}
\end{equation*}
where $\gamma^n_t$ denotes the distance of the two states in time $t$ and perturbed neuron $n$ and $\mean{\cdot}_{t, n}$ denotes the arithmetic average over time and all neurons.

The theory behind order and chaos is, of course, much more extensive and far beyond the scope of this paper. For more information, please refer to \cite{sprott2003chaos}.

\section{Information Theory Measures}
\label{sec:overview_information_theory_measures}

To gain an additional insight on what is happening inside a recurrent network, we will present two measures from the information-theoretical framework defined by \citeauthor{lizier2014framework} \cite{lizier2014framework}. The first measure is called \textit{active information storage (AIS)} and it denotes the average mutual information between the past states of a random process and its next state. Its definition is following:
\begin{equation*}
  \textstyle
  A_{X} = \lim_{k \to \infty} \sum_{X_n^{(k)},X_{n+1}}{p(X_n^{(k)},X_{n+1}) \log_2{\frac{p(X_n^{(k)},X_{n+1})}{p(X_n^{(k)})p(X_{n+1})}}} \,,
  \label{eq:active_information_storage}
\end{equation*}
where ${X_n^{(k)} = \left\{ X_{n}, X_{n-1}, ..., X_{n-k+1} \right\} }$ denotes the semi-infinite past of the process~$X_{n}$ and $p( \cdot )$ denotes the probability function. In the context of neural networks, the active information storage of a neuron measures how much does the neuron's history influence its future state. Self-links or transfers of information to other neurons and back are also considered.

The second measure is called \textit{transfer entropy (TE)}. It always regards two random processes, a \textit{source} and a \textit{destination}, between whom is the transfer entropy measured. It denotes the amount of information from the source which determines the value of the destination and was not already provided in the destination's history. In other words, it is the mutual information between the current state of the source process $Y$ and the next state of the destination process $X$ conditioned on the history of the destination process:
\begin{equation*}
  \textstyle
  T_{Y \to X} = \lim_{k \to \infty} \sum_{\omega_n} p(x_{n+1}, x^{(k)}_n, y_n) \log_2{\frac{p(x_{n+1}|x^{(k)}_n,y_n)}{p(x_{n+1}|x^{(k)}_n)}} \,,
  \label{eq:transfer_entropy}
\end{equation*}
In the context of neural networks, the transfer entropy measures how much does the current state of the source neuron influence the next state of the destination neuron. This measure was first introduced by \citeauthor{schreiber2000te} \cite{schreiber2000te} without the limit of $k \to \infty$, which was suggested later by \citeauthor{lizier2008te} \cite{lizier2008te}.

The aforementioned measures represent a universal tool for analyzing random processes. Please refer to \cite{lizier2014framework} for a unified overview.

\section{Neuroevolution}
\label{sec:overview_neuroevolution}

In this section, we will present two genetic algorithms specialized solely on neural networks. The first one is called \textit{NeuroEvolution of Augmenting Topologies (NEAT)} and the second one is its extension called \textit{Hypercube-based NeuroEvolution of Augmenting Topologies (HyperNEAT)}.

The NEAT algorithm was introduced by \citeauthor{stanley2002neat} in 2002 and it is still widely used. It has proven to be an efficient method to simultaneously evolve both, the weights of a neural network and the network's \textit{topology}. Please refer to the original paper by \citeauthor{stanley2002neat} \cite{stanley2002neat} for a thorough, yet succinct explanation.

The original NEAT algorithm evolves each connection independently, which is sufficient for small-scale neural networks. However, in the case of larger networks (i.e., hundreds of neurons), the number of connections is just too large for the evolution to succeed in a reasonable time. To overcome this issue, \citeauthor{gauci2007hyperneat} exploited the idea of \textit{indirect genetic encoding} and proposed an algorithm called HyperNEAT. This algorithm uses the original NEAT to evolve a population of neural networks called \textit{compositional pattern producing networks (CPPNs)}. The CPPNs are later presented with a user-defined set of neurons on a \textit{Cartesian plane}, called \textit{substrate}, and they are queried for the weight and the presence of each potential connection between all pairs of neurons. The substrate may be much larger than the CPPN itself and the CPPNs may thus compactly represent complex structures with repeating patterns and geometric regularities. For a more thorough description, please refer to the original paper by \citeauthor{gauci2007hyperneat} \cite{gauci2007hyperneat}.

\section{Performance Measures}
\label{sec:performance_measures}

To measure the computational power of a neural network, we have selected the following three benchmarks: \textit{Memory Capacity (MC)} \cite{jaeger2002short}, \textit{Nonlinear AutoRegressive Moving Average (NARMA)} \cite{boedecker2012information} and a novel measure called \textit{Negative Ratio (NR)}. They should assess the network's ability to store the data into a short-term memory (MC), operate with the memory (NARMA) and analyze it (NR).

The memory capacity (MC) task evaluates the maximum duration for which is the network capable of remembering its inputs. The evaluated network is driven by a single input sequence and predicts an infinite number of output sequences. The desired value of the $k$-th output sequence $y_k$ is an exact copy of the input sequence delayed by $k$ time steps. For each of the output sequences, the \textit{$k$-delay memory capacity} is calculated as the squared Pearson correlation coefficient between the predicted output and the desired output:
\begin{equation*}
  \textstyle
  \textit{MC}_k = \frac{\cov^2(y_k, o_k)} {\var(y_k) \var(o_k)} \,,
  \label{eq:k_memory_capacity}
\end{equation*}
where $\cov^2$ denotes the squared sample covariance, $\var$ denotes the sample variance, $y_k$ is the $k$-th desired output sequence, and $o_k$ is the $k$-th output sequence predicted by the network. The total memory capacity value is the sum of these $k$-delayed memory capacities:
\begin{equation*}
  \textstyle
  \textit{MC} = \sum_{k=1}^{\infty} \textit{MC}_k \,.
  \label{eq:memory_capacity}
\end{equation*}

During our experiments, we found this measure to be numerically unstable. When one of the output sequences has a very low variance (e.g., if the network always predicts a value close to 1.0), the $MC$ value is unpredictable and can go up to infinity. This may represent a problem especially in the case of evolutionary algorithms. Whenever an evolutionary algorithm detects such an instability of the fitness function, the instability is quickly exploited and the evolution converges to an undesired result. For this particular reason, we propose a numerically stable alternative of the memory capacity task called \textit{memory mean squared error (MMSE)}.

The evaluation of the MMSE is very similar to the evaluation of the MC. However, this time the network predicts only a finite number $N$ of output sequences. The desired value of the $k$-th output sequence is, again, the input sequence delayed by $k$ time steps. The final MMSE value is the \textit{normalized root mean squared error} of the predicted sequences with respect to the corresponding desired output sequences, as defined in the following formula:
\begin{equation*}
  \textstyle
  \textit{MMSE} = \sqrt{{\mean{(y^t_k - o^t_k )^2}_{t,k}}/{\var(y_0)}} \,,
  \label{eq:memory_mean_squared_error}
\end{equation*}
where $y^t_k$ and $o^t_k$ are the values of the $k$-th desired sequence and the $k$-th predicted sequence in time $t$, respectively. $y_0$ is the input sequence, and $\mean{\cdot}_{t,k}$ denotes the arithmetic average over time and all output sequences.

In Nonlinear AutoRegressive Moving Average (NARMA) task, the network is driven by a single input sequence and the task is to predict the following nonlinear combination of the past 30 inputs:
\begin{equation*}
  \textstyle
  y^{t+1} = 0.2y^t + 0.004y^t \sum_{i=0}^{29} y^{t-i} + 1.5x^{t-29}x^t + 0.001 \,,
  \label{eq:narma}
\end{equation*}
where $y^t$ and $x^t$ are the values of the desired output sequence and the input sequence in time $t$, respectively. The performance of this task is measured using the normalized root mean squared error:
\begin{equation*}
  \textstyle
  \textit{NARMA} = \sqrt{{\mean{(o^t - y^t)^2}_t}/{\var(y)}} \,,
  \label{eq:narma_nrmse}
\end{equation*}
where $o^t$ is the value of the predicted sequence in time $t$ and $\mean{\cdot}_{t}$ denotes the arithmetic average over time.

In the negative ratio (NR) task, the network shall estimate the ratio of negative numbers in the last $K$ inputs. NR is the only of the proposed measures, where the exact input values are not important and instead, the values shall be conditioned on a specific property. Formally, the desired output sequence is defined as following:
\begin{equation*}
  \textstyle
  y^t = \frac{1}{K} \sum_{i=0}^K \textrm{neg}(x^{t-i}) \,,
  \label{eq:negative_ratio_seq}
\end{equation*}
where $x^t$ and $y^t$ are the values of the input sequence and the desired output sequence in time $t$, respectively. $\textrm{neg}(x)$ is equal to $1$ iff $x \leq 0$ and $0$ otherwise. The performance of this task is measured using the normalized root mean squared error:
\begin{equation*}
  \textstyle
  \textit{NR} = \sqrt{{\mean{(o^t - y^t)^2}_t}/{\var(y)}} \,,
  \label{eq:nr_nrmse}
\end{equation*}
where $o^t$ is the value of the predicted sequence in time $t$.

\section{Random Echo State Networks}
\label{sec:esn_at_edge_of_chaos}

In this section, the results of the related works by \citeauthor{bertschinger2004real} \cite{bertschinger2004real} and \citeauthor{boedecker2012information} \cite{boedecker2012information} are replicated. Both papers suggest that the computational power of randomly generated echo state networks is maximized on the transition between order and chaos. Furthermore, \citeauthor{boedecker2012information} suggest that there is a peak of active information storage and transfer entropy right on the edge of chaos.

\subsection{Experimental Settings}
\label{ssec:esn_settings}

The experiment is conducted by generating a large set of random echo state networks of different parameters. All the evaluated echo state networks have 151 neurons (including the input neuron) and use hyperbolic tangent transfer function with no bias. The weights from the input neuron to all the other neurons are drawn uniformly from the range of $[-0.1; 0.1]$. All the other weights are drawn from a normal distribution with zero mean and variance~$\sigma$. The $\sigma$ values are chosen so that $\log ( \sigma )$ is in the range of $[-3.7; -0.8]$ increasing its value by $0.02$. For each $\sigma$, we generate and evaluate 10 random networks. The input sequences for the MC task are drawn uniformly from the range of $[-1; 1]$. For the NARMA task, the range is $[0; 0.5]$. The length of the input sequences is 3000 time steps. The first 1000 time steps are used to stabilize the network, i.e., the network is driven by the sequence, but its outputs are discarded. The next 1000 time steps are used to train the linear coefficients and the last 1000 time steps are used to evaluate the network's performance. The MC is evaluated only up to the delay limit of 300 time steps.

To evaluate AIS and TE, an input sequence of length 3000 time steps is generated uniformly from the range of $[0; 0.5]$. The first 1000 steps are used to stabilize the network and the remaining 2000 time steps are used to calculate the measures. A history of size 2 is used for both the measures, instead of infinity. The AIS is averaged over all the neurons in the network and the TE is averaged over all the nonzero connections.

The experiment consumed approximately 120 CPU days.

\subsection{Results}
\label{ssec:esn_performance_results}

\begin{figure*}
  \centering
  \begin{subfigure}[t]{0.32\textwidth}
    \centering
    \includegraphics[width=\textwidth]{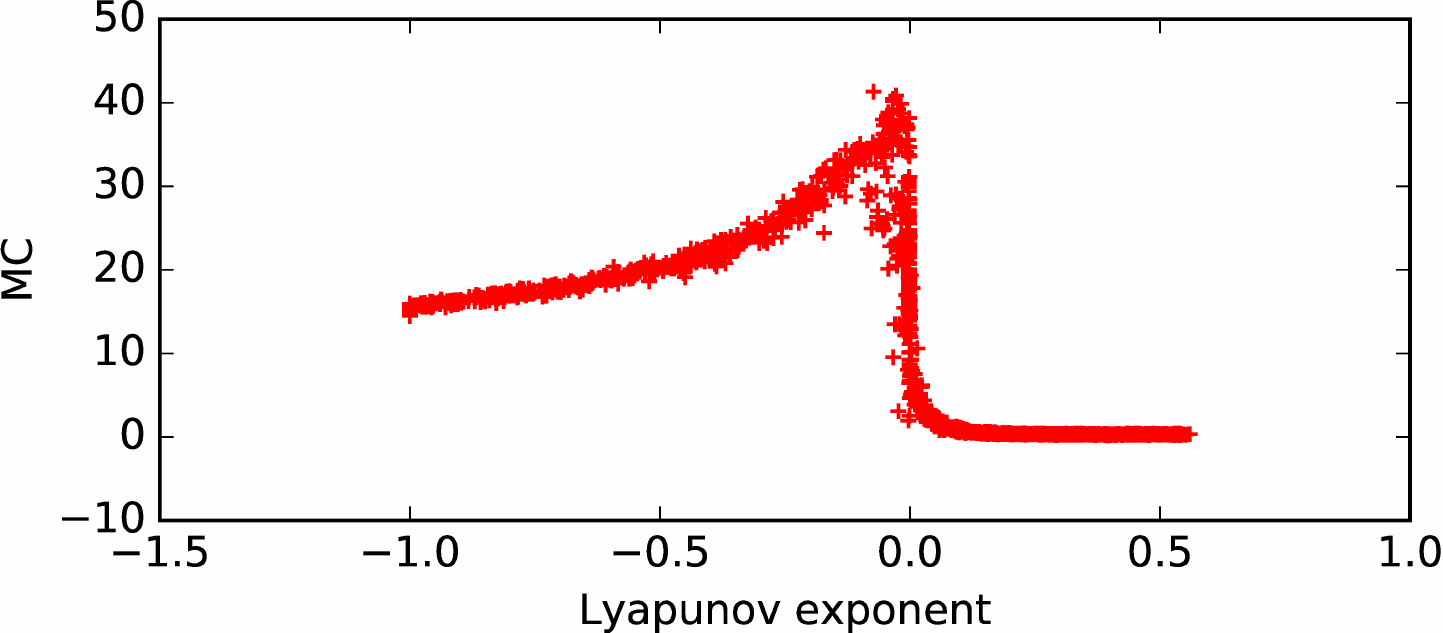}
    \caption{MC versus $\lambda$. Higher is better.}
    \label{fig:esn_full_lyap_vs_memory}
  \end{subfigure}
  \hfill
  \begin{subfigure}[t]{0.32\textwidth}
    \centering
    \includegraphics[width=\textwidth]{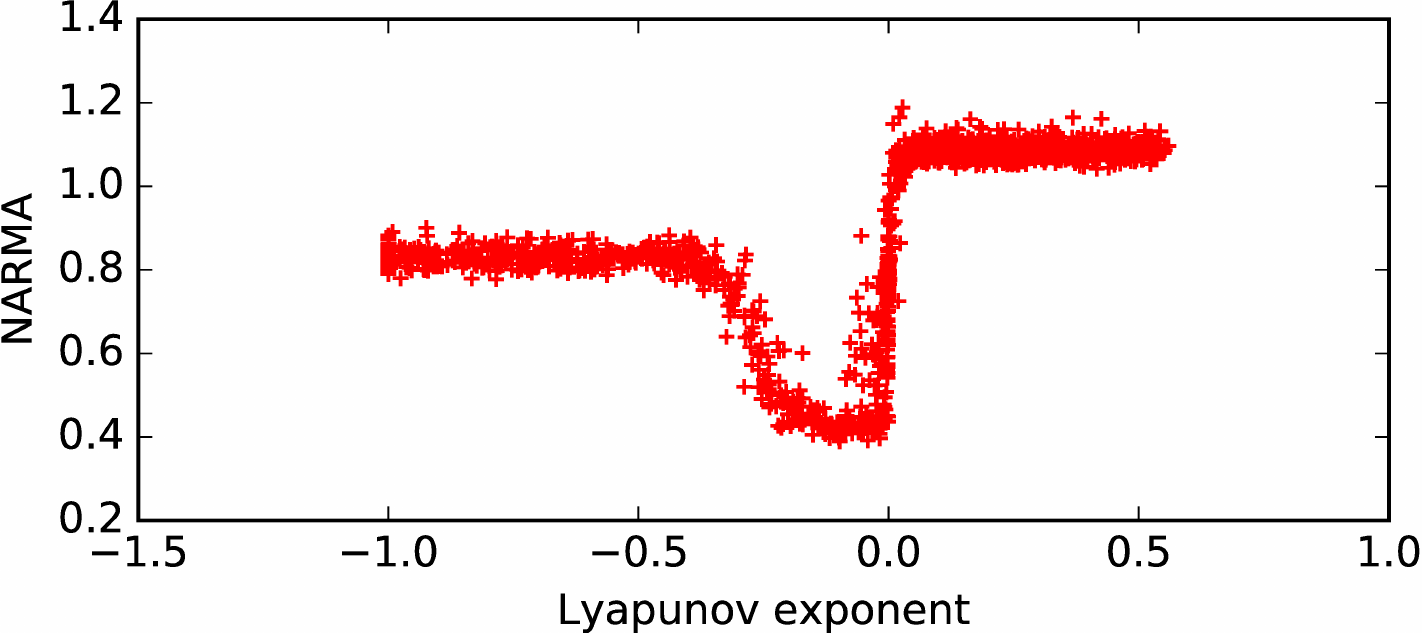}
    \caption{NARMA versus $\lambda$. Lower is better.}
    \label{fig:esn_full_lyap_vs_narma}
  \end{subfigure}
  \hfill
  \begin{subfigure}[t]{0.32\textwidth}
    \centering
    \includegraphics[width=\textwidth]{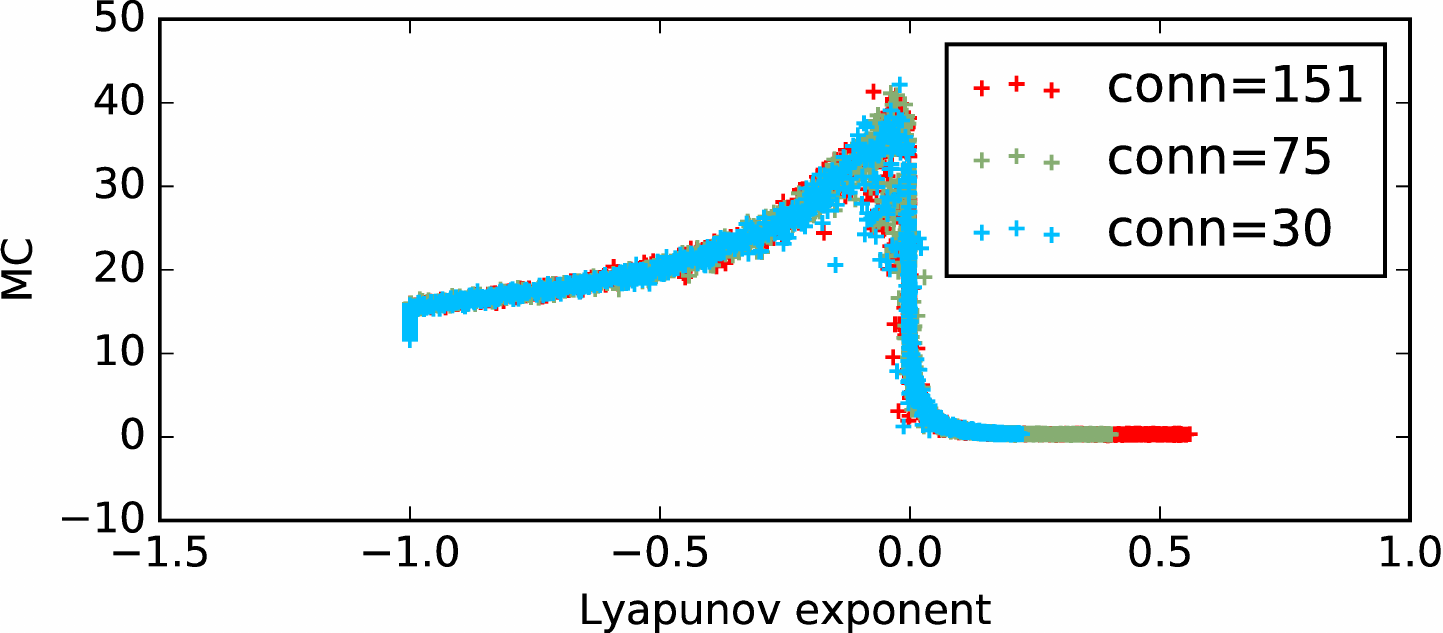}
    \caption{MC versus $\lambda$. Comparison of fully connected networks and networks randomly pruned to 75 and 30 connections per neuron.}
    \label{fig:esn_partial_lyap_vs_mc}
  \end{subfigure}
  \caption{Plots of various measures versus $\lambda$. Each cross represents a single random echo state network.}
\end{figure*}

The memory capacity (MC) against $\lambda$ is plotted in Figure~\ref{fig:esn_full_lyap_vs_memory}. The MC increases until the edge of chaos, where it reaches its maximum value. After the performance peak, there is a sharp drop, which suggests that in the chaotic regime, even if it is very close to the transition, no data survive the surrounding noise for long.

Our results of the MC task actually differ from the original paper by \citeauthor{boedecker2012information} \cite{boedecker2012information}, where the MC only rarely reached a value higher than 10. On the other hand, our results are in accordance with the paper by \citeauthor{barancok2014memory} \cite{barancok2014memory} who investigated the effect of structured input sequences (i.e., sequences which are not purely random) on the MC task.

The NARMA error against $\lambda$ is plotted in Figure~\ref{fig:esn_full_lyap_vs_narma}. The error is decreasing until the edge of chaos, where it reaches its minima. After the transition to the chaotic regime, the error sharply raises to the same value as if the network's output would be absolutely random. This observation supports the idea that in the chaotic regime, the network output resembles a random white noise.

We have evaluated also the MMSE and NR tasks and the results are very similar. It seems that for all the evaluated tasks, the performance is indeed maximized on the edge of chaos. A rigorous reason for this behaviour remains an open question.

Let us analyze the effects of randomly removing the majority of the connections between the neurons. According to our simulations, such a restriction of the connectivity does not significantly influence the network's performance. Figure~\ref{fig:esn_partial_lyap_vs_mc} demonstrates this phenomena on the MC task and a similar pattern appears on the other evaluated performance tasks as well. It should be noted that restricting the number of connections while keeping the same weights makes the network dynamics more ordered (as stated by, e.g., \citeauthor{bertschinger2004real} \cite{bertschinger2004real}).

\begin{figure*}
  \centering
  \begin{subfigure}[t]{0.32\textwidth}
    \centering
    \includegraphics[width=\textwidth]{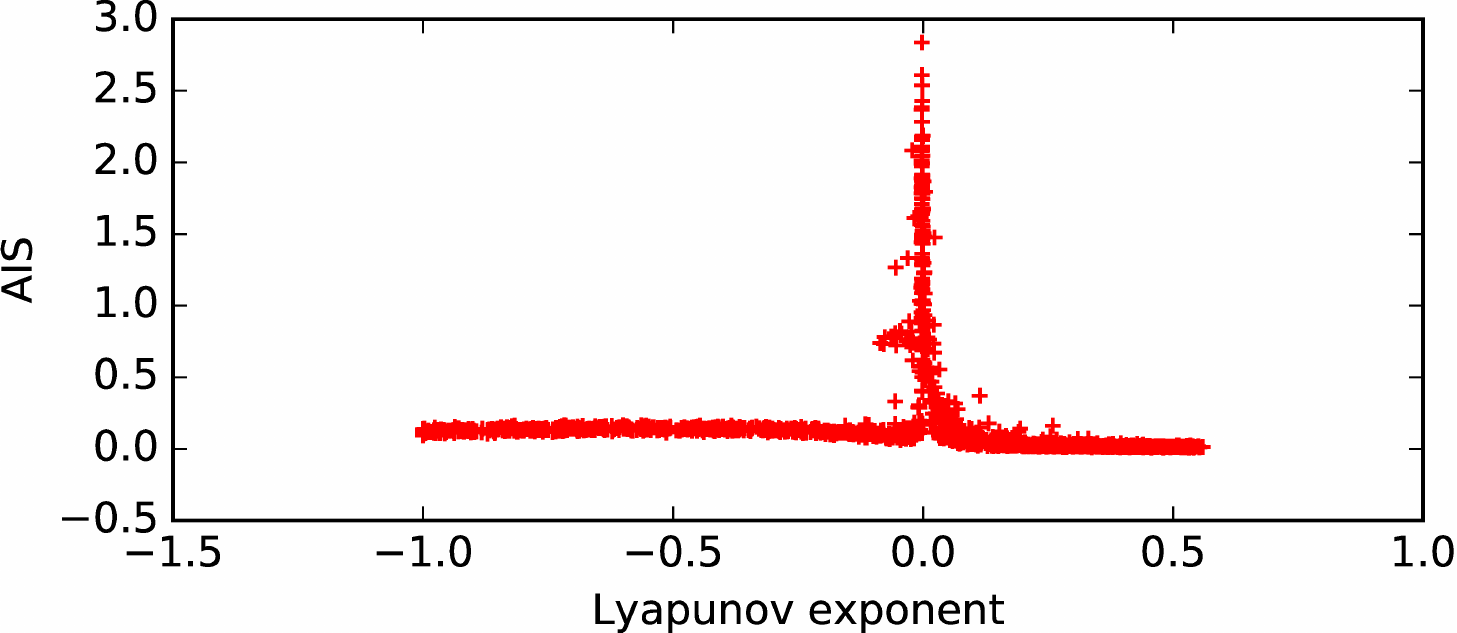}
    \caption{AIS versus $\lambda$.}
    \label{fig:esn_full_lyap_vs_ais}
  \end{subfigure}
  \hfill
  \begin{subfigure}[t]{0.32\textwidth}
    \centering
    \includegraphics[width=\textwidth]{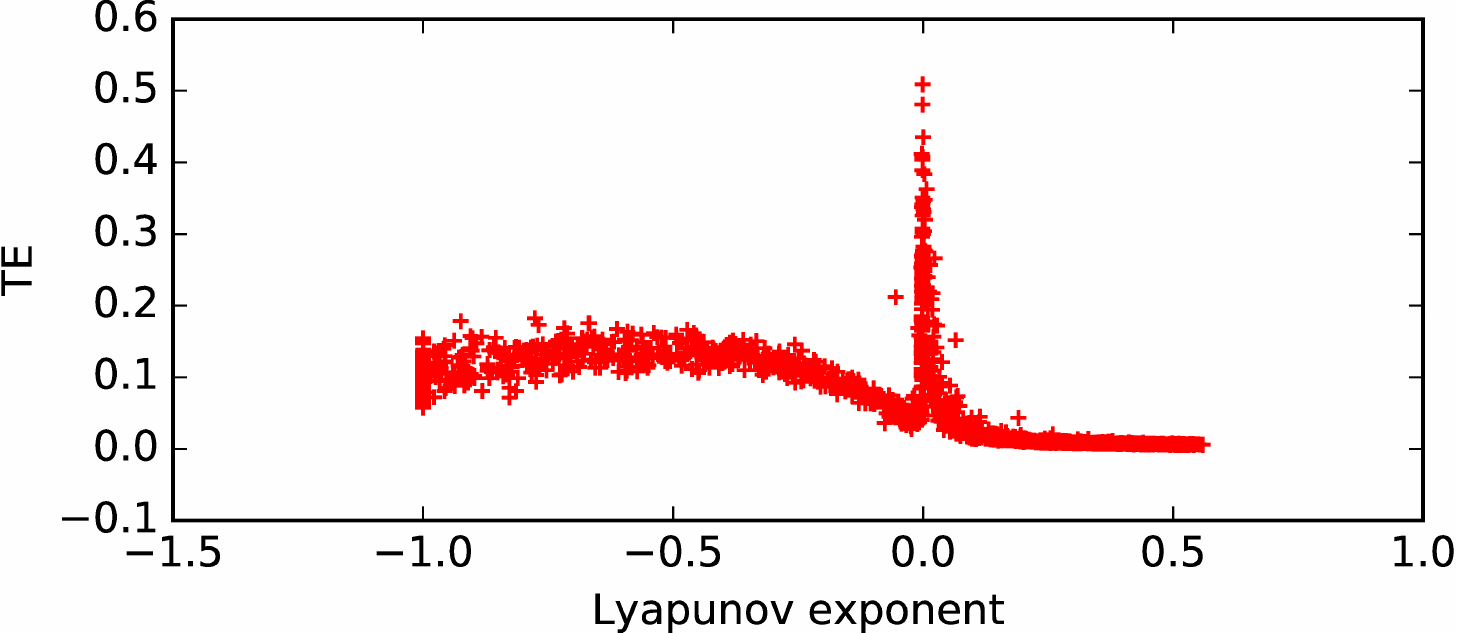}
    \caption{TE versus $\lambda$.}
    \label{fig:esn_full_lyap_vs_te}
  \end{subfigure}
  \hfill
  \begin{subfigure}[t]{0.32\textwidth}
    \centering
    \includegraphics[width=\textwidth]{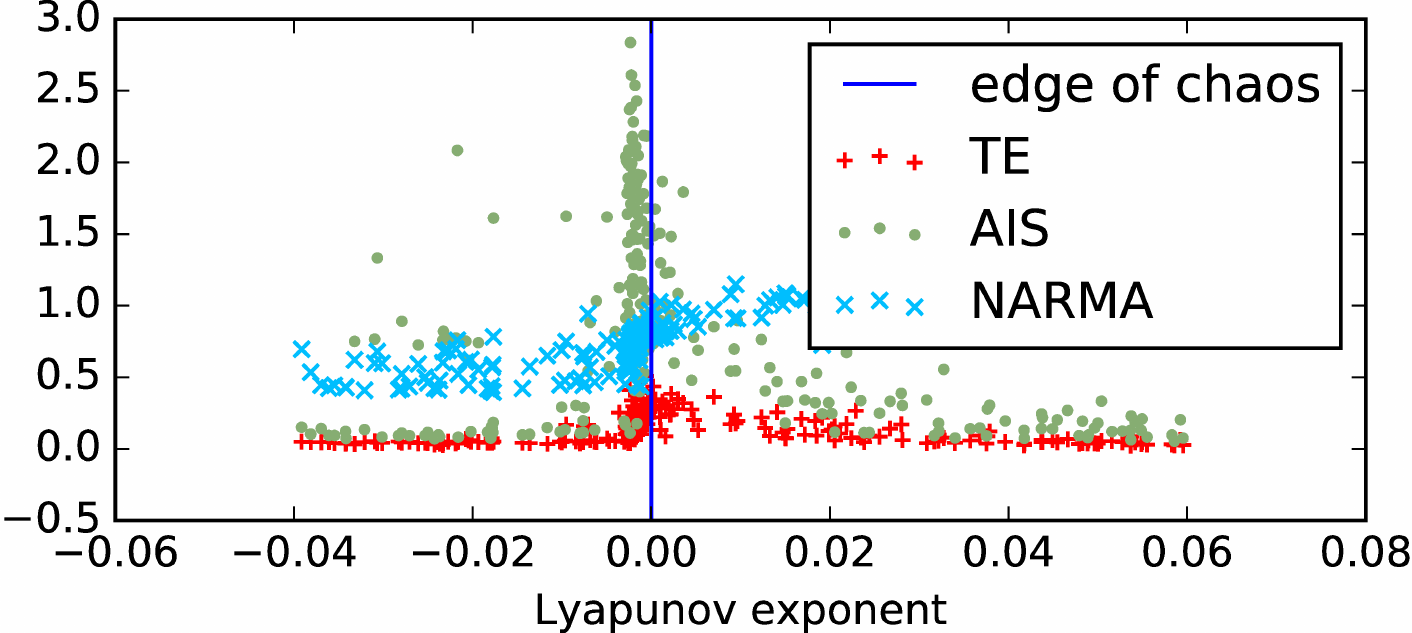}
    \caption{AIS, TE and NARMA versus $\lambda$.}
    \label{fig:esn_full_lyap_vs_te_ais_narma}
  \end{subfigure}
  \caption{Information-theory measures versus $\lambda$. Each point represents a single random echo state network.}
\end{figure*}

\citeauthor{boedecker2012information} \cite{boedecker2012information} provide an additional insight on what is happening within a network on the edge of chaos. In the paper, the AIS and TE are measured, relative to $\lambda$. The results are replicated in Figure~\ref{fig:esn_full_lyap_vs_ais} and Figure~\ref{fig:esn_full_lyap_vs_te}. Both of the measured entropies slowly decrease through the ordered regime until the edge of chaos, where they form a high sharp peak. In the chaotic regime, their value drops and remains at its lowest level. There does not seem to be a direct correlation between the network's performance and the AIS or the TE. Nevertheless, the high peak leads to the belief that there is an unexpected irregularity on the transition between the order and chaos.

In the original paper by \citeauthor{boedecker2012information} \cite{boedecker2012information}, the AIS and TE were plotted separately from the performance measures. Let us, instead, draw the entropies and the performance measures to a single plot focused tightly on the edge of chaos (Figure~\ref{fig:esn_full_lyap_vs_te_ais_narma}). After a careful analysis of the figure, it can be seen that the best performance on the NARMA task ends immediately \textit{before} the peaks of AIS and TE. When the entropies increase, the NARMA error increases with them. The reason may be that when the entropies reach a critical threshold, the network performance is impaired.

\section{Evolved Echo State Networks}
\label{sec:evolution_of_esn}

In this section, we are going to evolve the recurrent part of echo state networks via HyperNEAT algorithm. To avoid overfitting to one of the given tasks, the evolution in our experiment is instructed to maximize the performance on the MMSE and the NARMA tasks simultaneously. We believe that these two tasks require contradictory properties of the evolved network, which may reduce overfitting by balancing between memory capacity and computational performance. Furthermore, the NR task is hidden to the evolution and it is instead used to \textit{validate} the performance of the evolved network on tasks never seen before. The MC task is not evaluated at all because of its numerical instability discussed in Section~\ref{sec:performance_measures}. The MMSE task is used to asses the network's short-term memory instead.

\subsection{Experimental Settings}
\label{ssec:evo_settings}

Five runs of evolution are executed, each of which evolves a population of 150 CPPNs for 2000 generations. The substrate used in our experiment is a ``\textit{golden angle spiral}'', in which the coordinate $[X_k; Y_k]$ of the $k$-th neuron is defined as $X_k = \sqrt{{k}/{N}} \cos(k \varphi + \omega)$ and $Y_k = \sqrt{{k}/{N}} \sin(k \varphi + \omega)$, where $N$ is the number of neurons, $\varphi$ is the ``golden angle'' equal to $\pi (3 - \sqrt 5 )$, and $\omega$ is the rotation angle (i.e., the phase) of the whole spiral. The rotation angle $\omega$ is generated randomly for each of the five evolutionary runs.

The size of the substrate is 151 neurons. A single input neuron, whose activation always corresponds to the current input value, is placed in the centre of the substrate, on coordinate $[0; 0]$. When using a CPPN to build a connection between two neurons, the CPPN is fed by the distance between the neurons in addition to their coordinates. The networks generated on the substrate use hyperbolic tangent transfer function with no bias. Neurons disconnected from the input are not considered for $\lambda$, AIS and TE measures.

The fitness function of a CPPN is defined as $2 - \textit{MMSE}(x) - \textit{NARMA}(x)$, where $x$ is the network generated by the CPPN on the aforementioned substrate. The fitness value is evaluated three times and the results are averaged. For the first 15 generations of a life of a genome, its fitness is boosted by 10\%. The number of species is kept between 5 and 10. The difference between two genomes is defined as ${F}/{N} + {\overline{W}}/{2}$, where $N$ is the number of genes in the larger genome, $F$ is the number of non-matching genes, and $\overline{W}$ is the average weight difference of matching genes. The difference threshold $\delta_t$ for creating a new species begins at $2.0$ and may be dynamically increased or decreased in each generation by $0.3$.

Only the fittest 25\% genomes of a species are allowed to reproduce. The elitism is set to 5\%. If a species does not improve its fitness for 20 generations, it is forbidden to reproduce. To select the best genomes, \textit{tournament selection} of size four is used on the top 25\% of the genomes in the species. Overall crossover probability is 70\%. There is only a 0.01\% chance of interspecies mating.

Overall mutation probability is 15\%. If the mutation occurs, the chance of adding a new neuron is 1\%, the chance of adding a new connection is 8\%, the chance of removing a connection is 2\%, and each weight has a 90\% chance of being perturbed by a uniformly distributed random value from the range of $[-0.2; 0.2]$. The probability of mutating the bias of a neuron is 1\% and the mutation process is the same as for the weight mutation.

Additionally, if the mutation occurs, the transfer function of each neuron might mutate as well. The probability of mutating the transfer function is 3\% and in such a case, the function is replaced by one of the following functions: hyperbolic tangent, sine, signed step, signed gaussian, and linear transfer function.

We have evaluated a few more different configurations and the results seemed to be insensitive to small parameter changes. The experiment consumed approximately 80 CPU days.

\subsection{Results}
\label{ssec:evo_results}

\begin{figure*}
  \centering
  \begin{subfigure}[t]{0.32\textwidth}
    \centering
    \includegraphics[width=\textwidth]{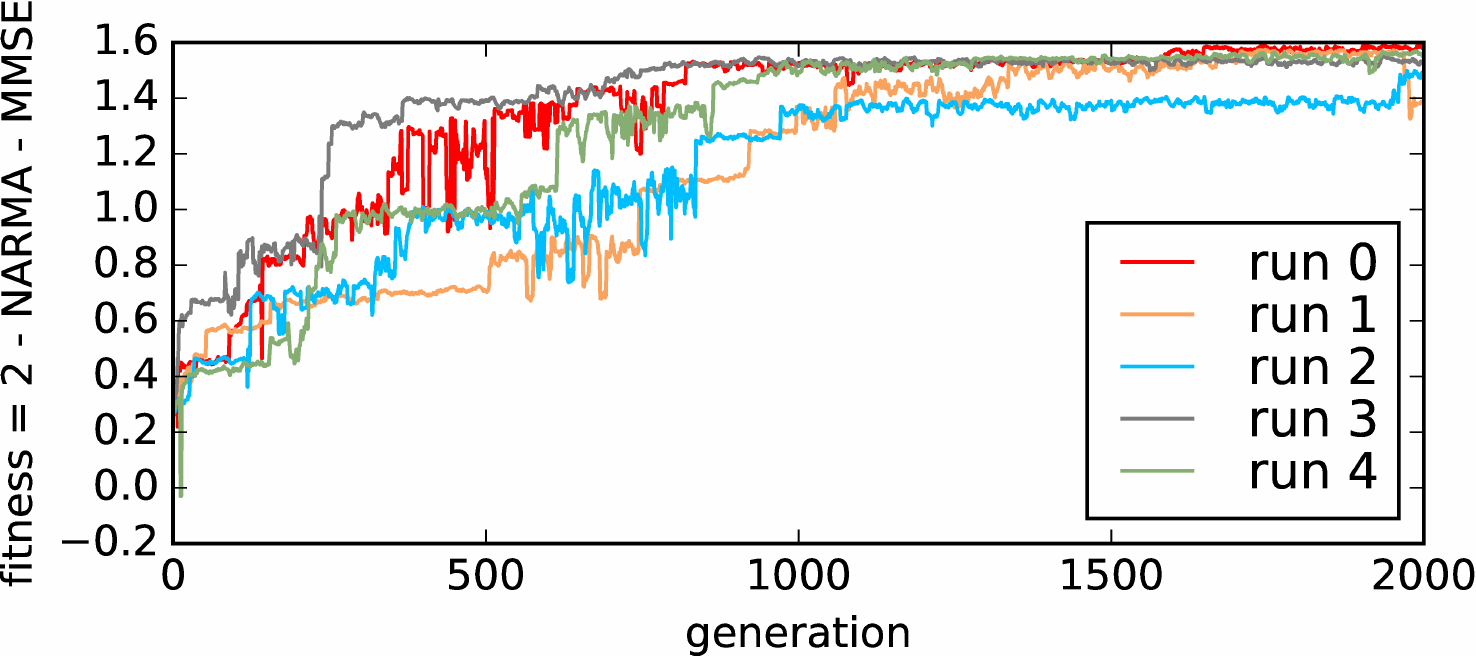}
    \caption{The evolution of the fitness.}
    \label{fig:evolution_fitness}
  \end{subfigure}
  \hfill
  \begin{subfigure}[t]{0.32\textwidth}
    \centering
    \includegraphics[width=\textwidth]{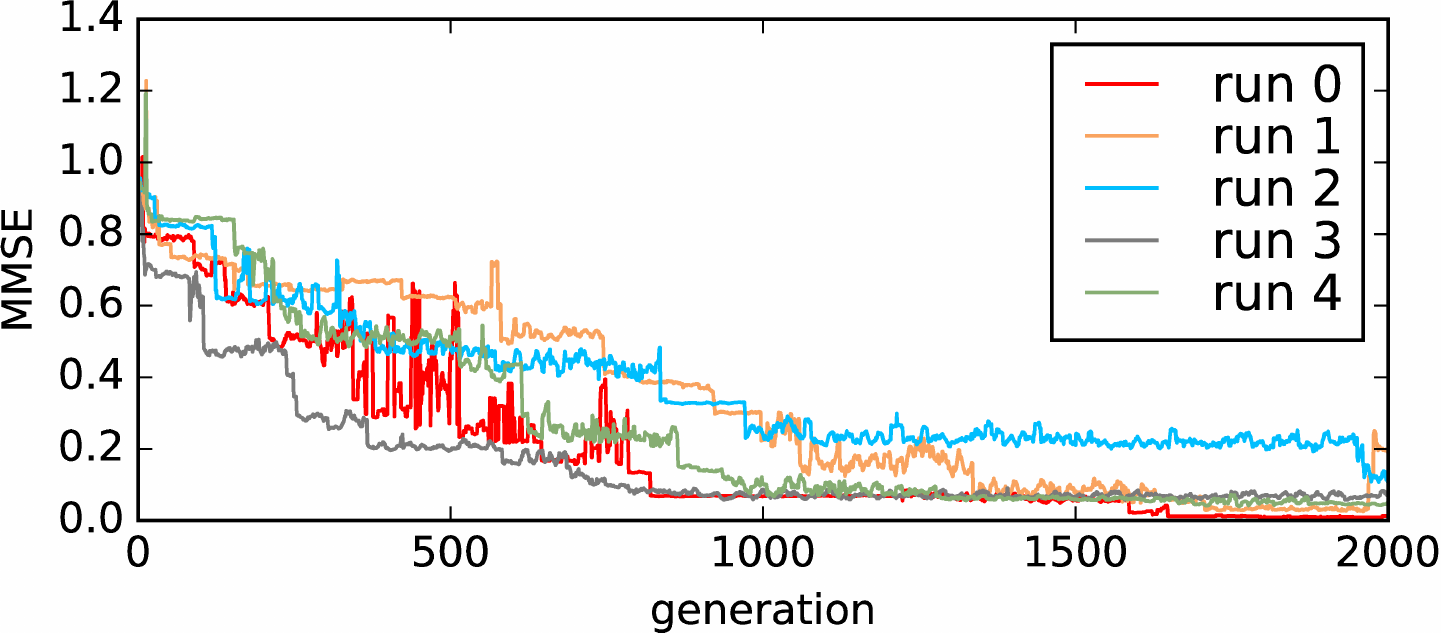}
    \caption{The evolution of MMSE error.}
    \label{fig:evolution_mmse}
  \end{subfigure}
  \hfill
  \begin{subfigure}[t]{0.32\textwidth}
    \centering
    \includegraphics[width=\textwidth]{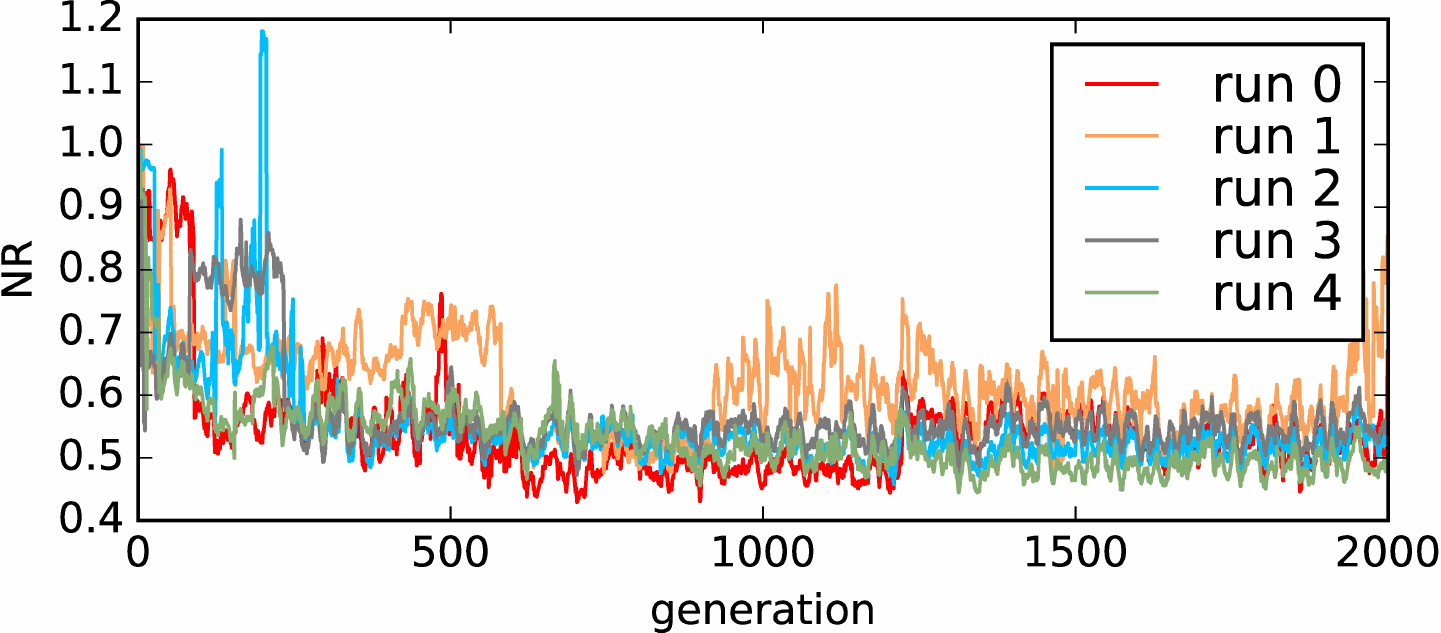}
    \caption{The evolution of NR error.}
    \label{fig:evolution_nr}
  \end{subfigure}
  \caption{The evolution of the best individual in a generation.}
\end{figure*}

The evolution of the fitness value ($2 - \textit{MMSE} - \textit{NARMA}$) is plotted in Figure~\ref{fig:evolution_fitness}. All the evolutionary runs converged stably to a similar value. The improving evolution of MMSE is depicted in Figure~\ref{fig:evolution_mmse} and NARMA manifested similar results (not depicted). Since MMSE and NARMA form the fitness function, it is no surprise they consistently improve throughout the evolution. However, also the NR task (Figure~\ref{fig:evolution_nr}), that was not optimized directly, slightly improved and stabilized, suggesting that the evolved network is not overfitted to the fitness function. 

\begin{table*}
  \centering
  \setlength{\tabcolsep}{5pt}
  \begin{tabular}{lrrrr}
    \toprule
          & MMSE & NARMA & NR & Lyap. exp. \\
    \midrule
    evolved &  0.0094 ($\pm$ 0.0009 std) &  0.4119 ($\pm$ 0.0297 std) &  0.5403 ($\pm$ 0.0688 std) & -0.2652 ($\pm$ 0.0000 std) \\
    random  &  0.2625 ($\pm$ 0.0098 std) &  0.4150 ($\pm$ 0.0171 std) &  0.5728 ($\pm$ 0.0730 std) & -0.0280 ($\pm$ 0.0000 std) \\
    p-value &  $\boldsymbol<\mathbf{0.01}$  &  0.2066 &  $\boldsymbol<\mathbf{0.01}$ &  \\
    \bottomrule
  \end{tabular}
  \caption{The comparison of the best evolved network and the best random echo state network on 50 full evaluation cycles (stabilize, train, evaluate). The p-values for the hypothesis that the evolved network outperforms the random network are calculated using one-tailed one-sample t-test.}
  \label{tab:best_random_esn_vs_evo}
\end{table*}

To compare the evolved networks with the orginal random echo state networks, the best representatives of both categories are selected. The selection criteria is the average of ten evaluations of the fitness function. The two best representatives are statistically compared in Table~\ref{tab:best_random_esn_vs_evo}. The performance on the NARMA task is similar for both the random and the evolved network. However, on the MMSE task, the evolved network outperforms the random network by more than one order of magnitude. On the NR task, which has not been explicitly optimized by any of the two approaches, the evolved network also performs significantly better.

\begin{figure}
  \centering
  \includegraphics[width=0.9\linewidth]{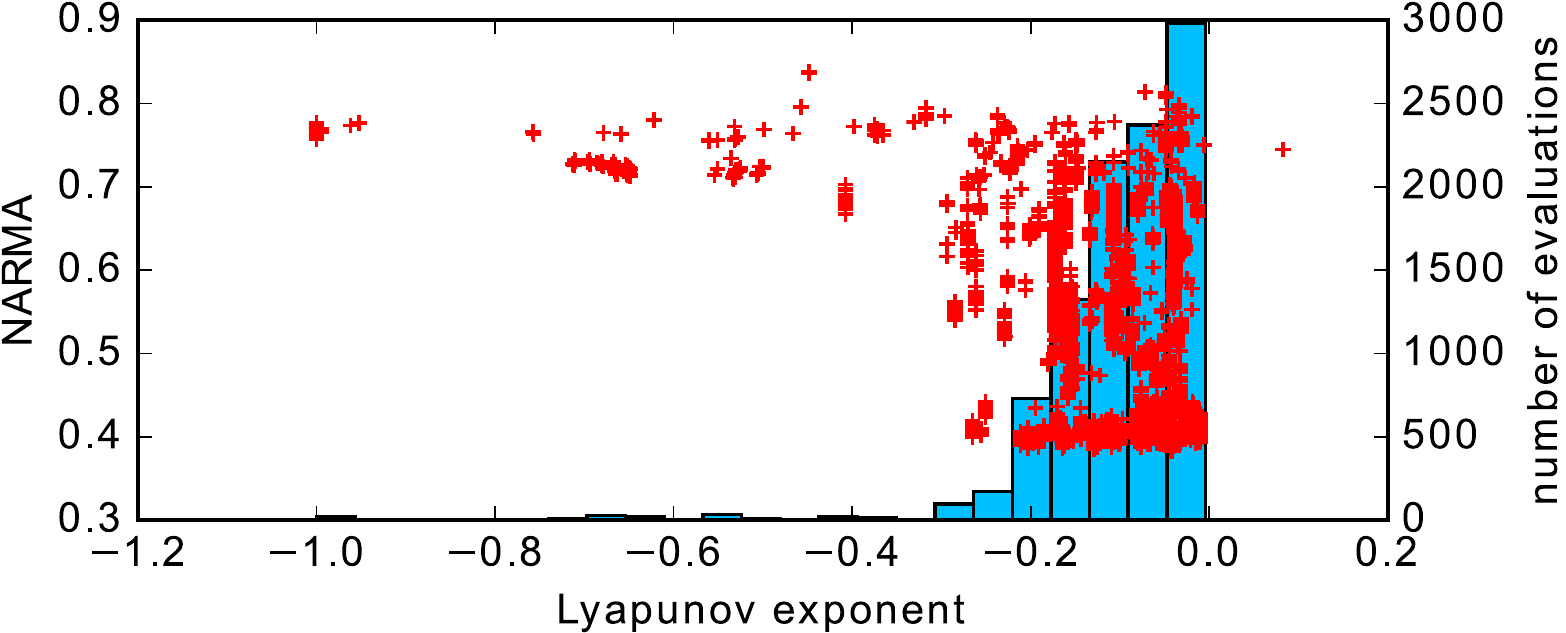}
  \caption{Statistics over all the five evolutionary runs. Red - NARMA versus $\lambda$; each cross denotes a single evaluation of an evolved network. Blue - the histogram of $\lambda$'s.}
  \label{fig:evolution_lyap_vs_narma}
\end{figure}

Now, we will address the question whether the evolution has any relation to the edge of chaos. Figure~\ref{fig:evolution_lyap_vs_narma} plots NARMA versus $\lambda$ and the histogram of $\lambda$'s encountered during any of the five runs of the evolution. Only a very few networks have strongly ordered dynamics and their performance is rather poor. The vast majority of evolved networks is concentrated on the ordered side of the edge of chaos, in the range of $[-0.3; 0]$ of $\lambda$. It is clear that the evolution avoided chaotic regime at all costs. The results might suggest that the ordered side of the edge of chaos is a part of the search space favoured by evolutionary algorithms. One explanation may be, that if the evolution is completely unable to solve the given task, it may generate a network more or less randomly. If this random network is on the edge of chaos, its performance is still better compared to the performance of an overly ordered or a chaotic network.

\subsection{Topology}
\label{ssec:evo_topology}

\begin{figure}
  \centering
  \begin{subfigure}{0.4\linewidth}
    \centering
    \includegraphics[width=\textwidth]{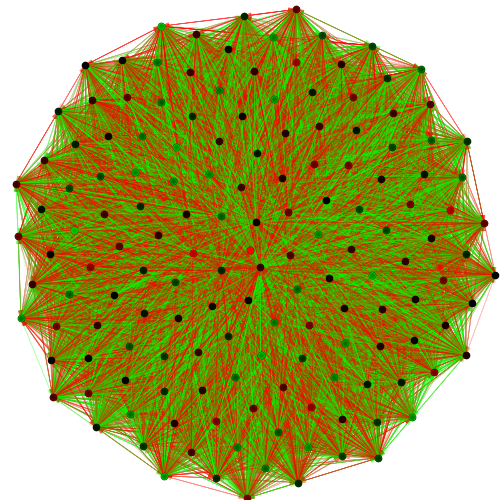}
    \label{fig:best_random_esn}
  \end{subfigure}
  \hfill
  \begin{subfigure}{0.4\linewidth}
    \centering
    \includegraphics[width=\textwidth]{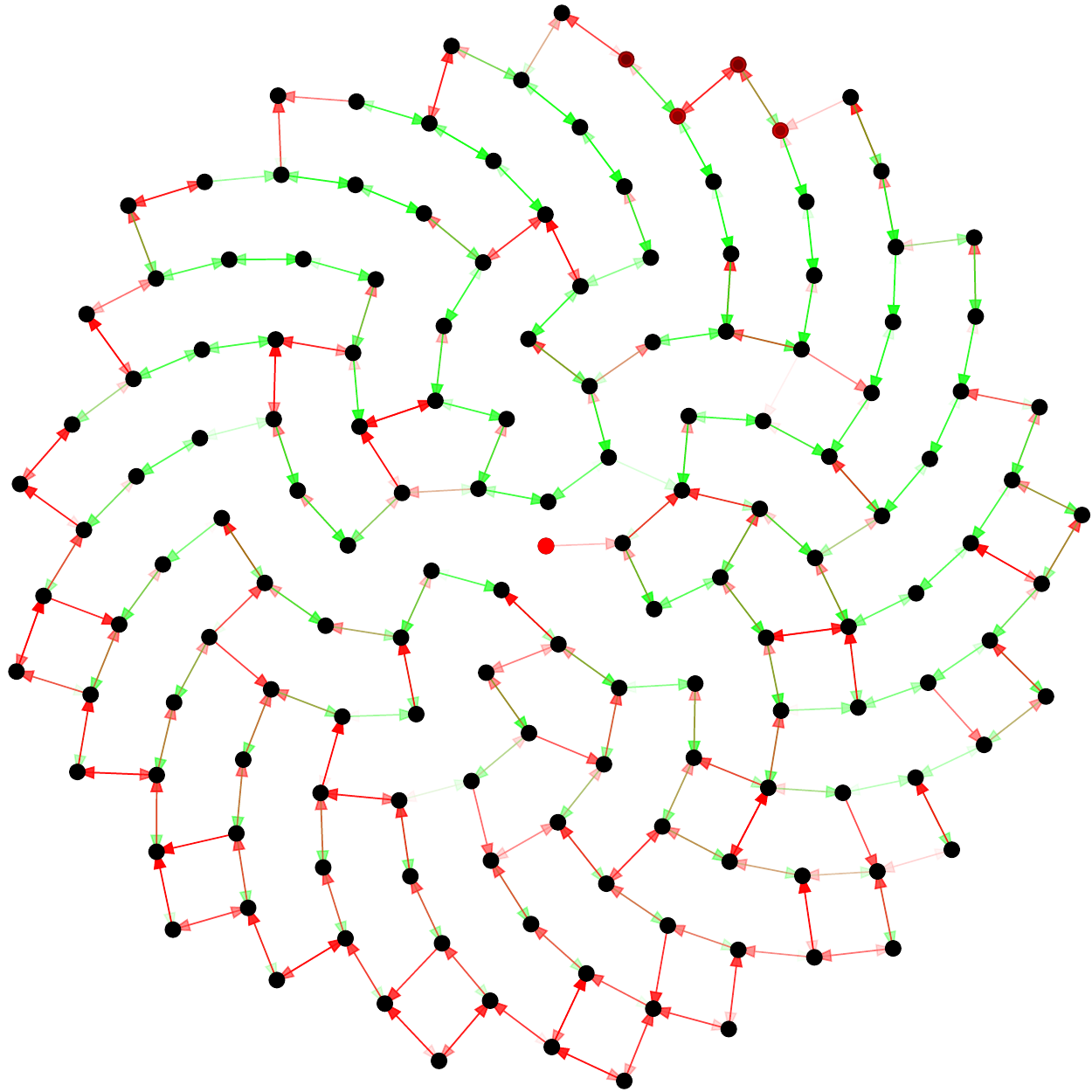}
    \label{fig:best_evolved_esn}
  \end{subfigure}
  \caption{Visualization of the best random (left) and the best evolved (right) echo state networks. Green color represents positive weights and red color represents negative weights.}
  \label{fig:best_random_evolved_esn}
\end{figure}

\begin{figure}
  \centering
  \setlength\tabcolsep{3pt}
  \begin{tabular}{cccc}
    \includegraphics[width=0.22\linewidth]{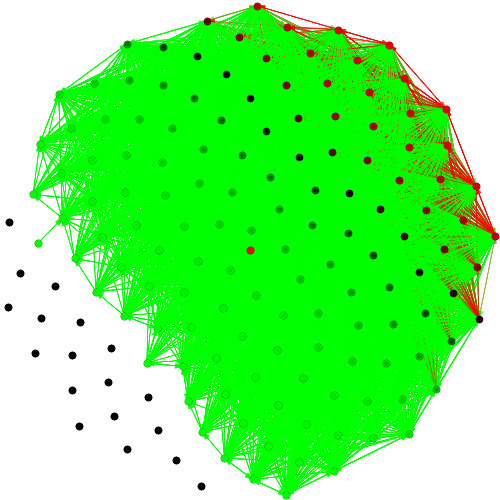} & \includegraphics[width=0.22\linewidth]{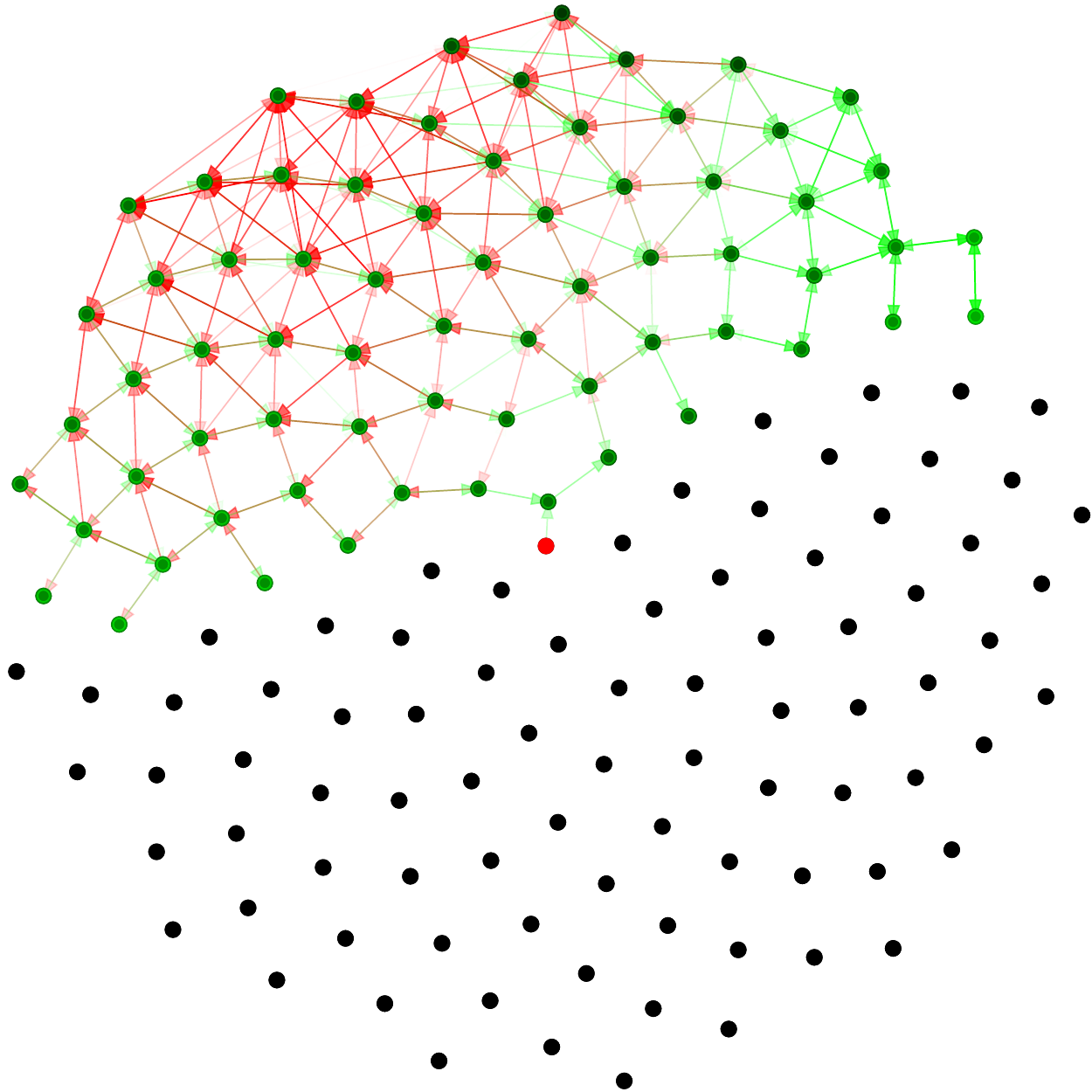} & \includegraphics[width=0.22\linewidth]{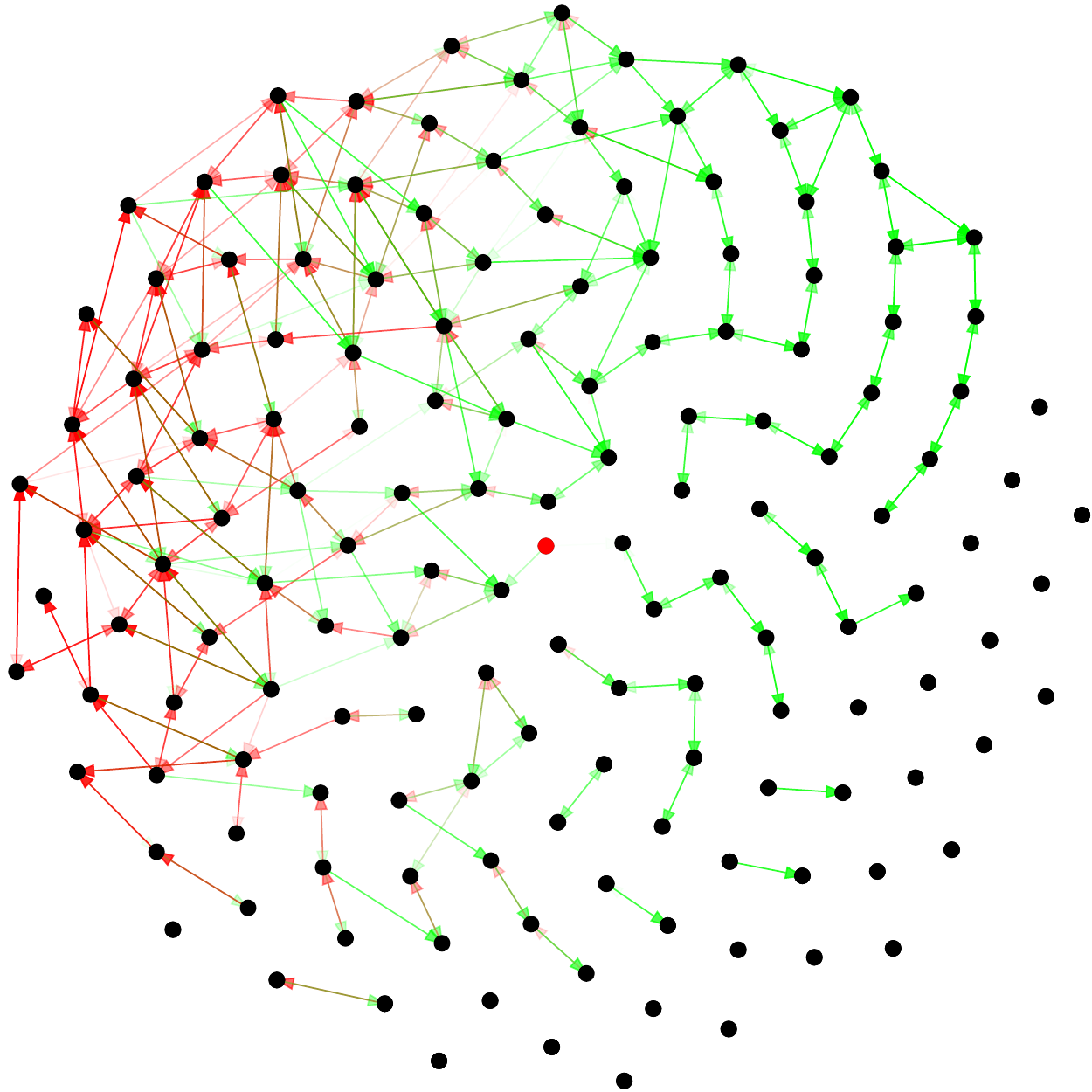} & \includegraphics[width=0.22\linewidth]{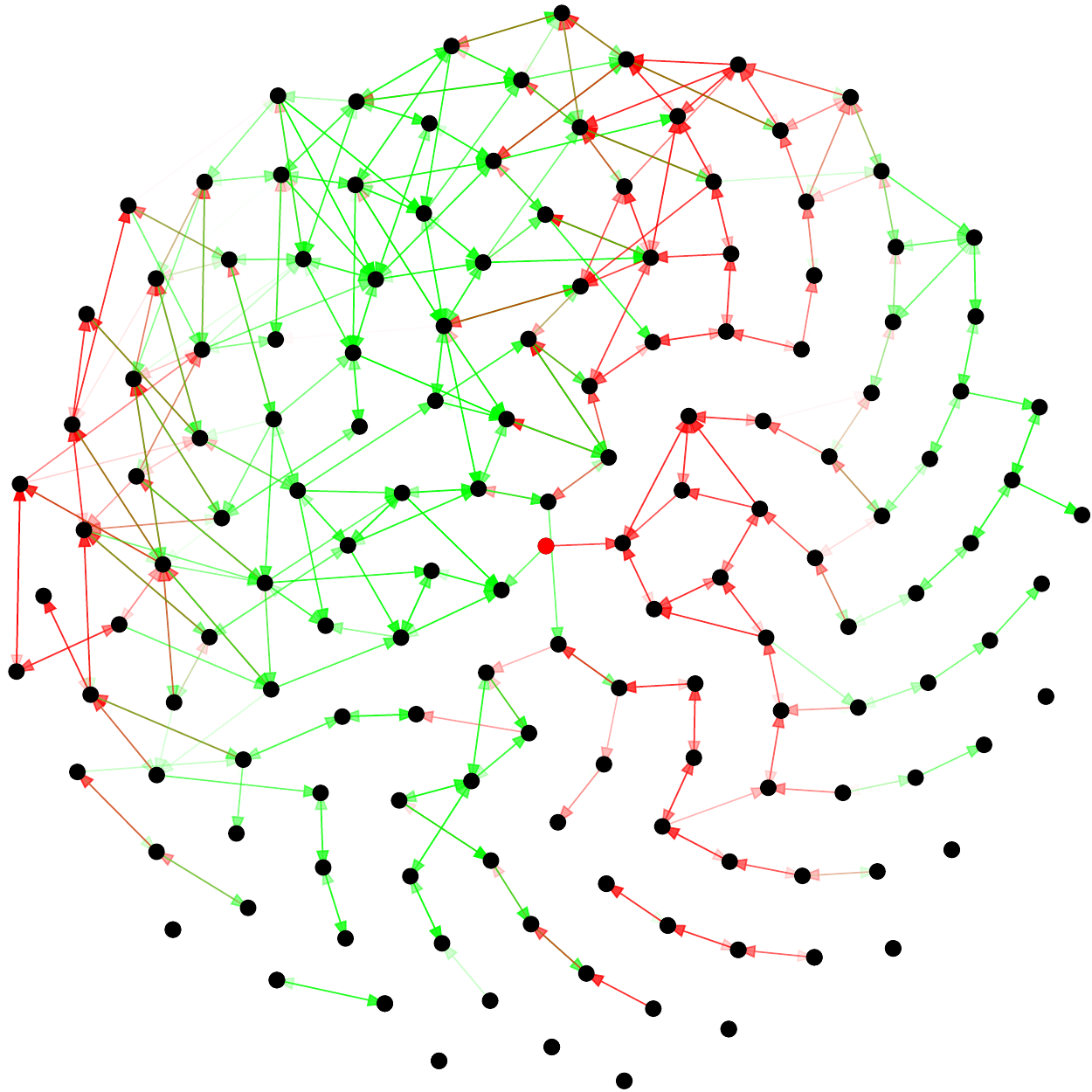} \\ 
    \includegraphics[width=0.22\linewidth]{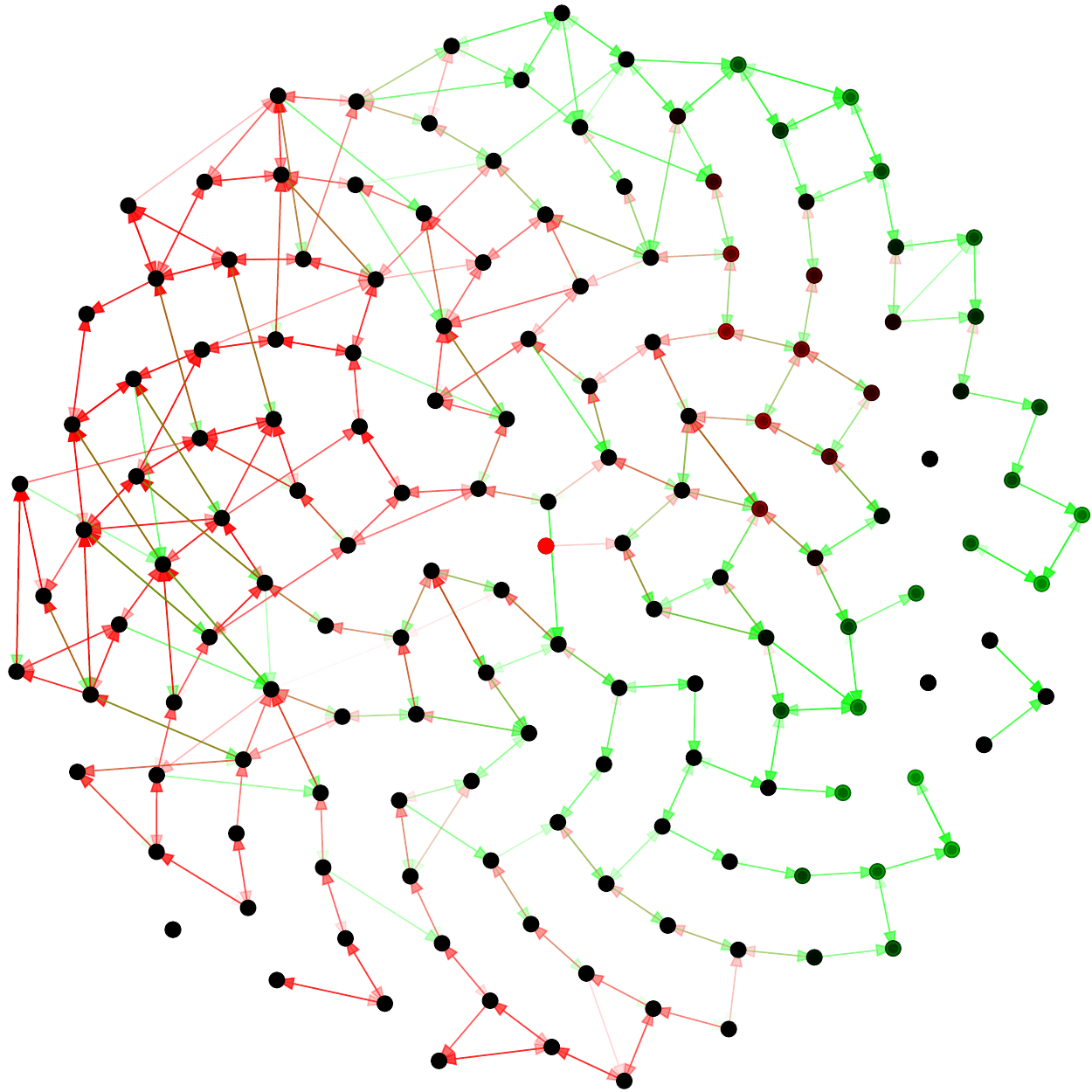} & \includegraphics[width=0.22\linewidth]{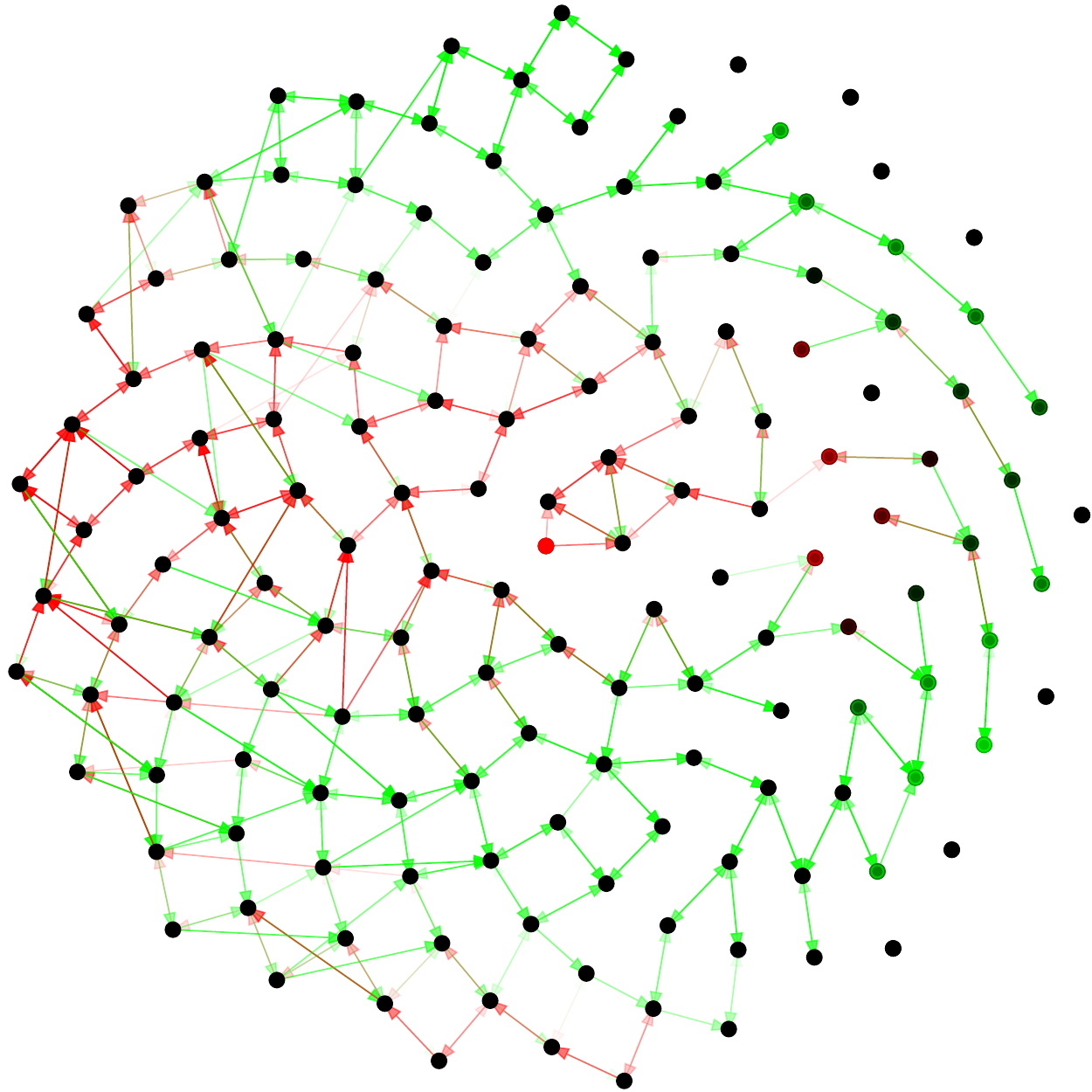} & \includegraphics[width=0.22\linewidth]{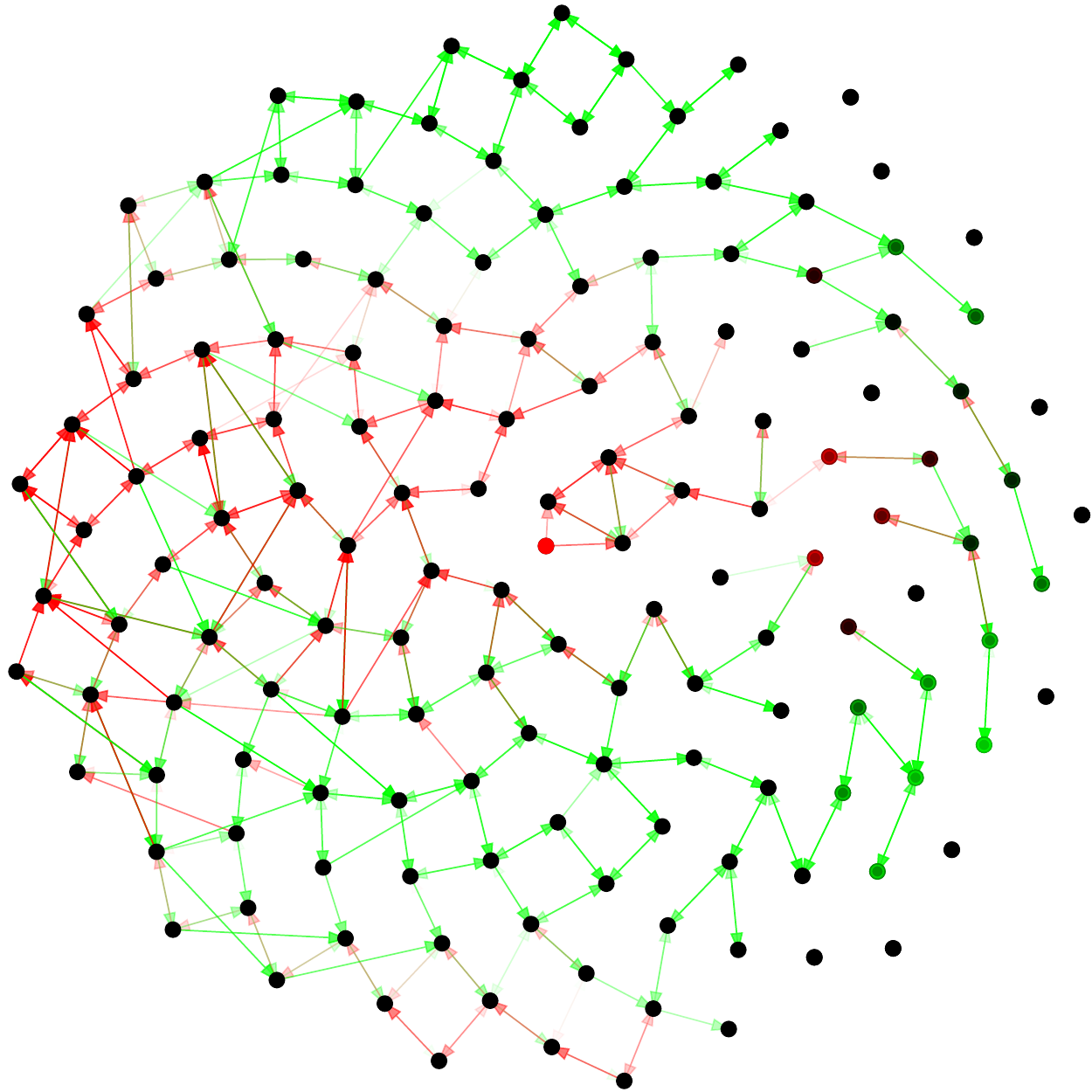} & \includegraphics[width=0.22\linewidth]{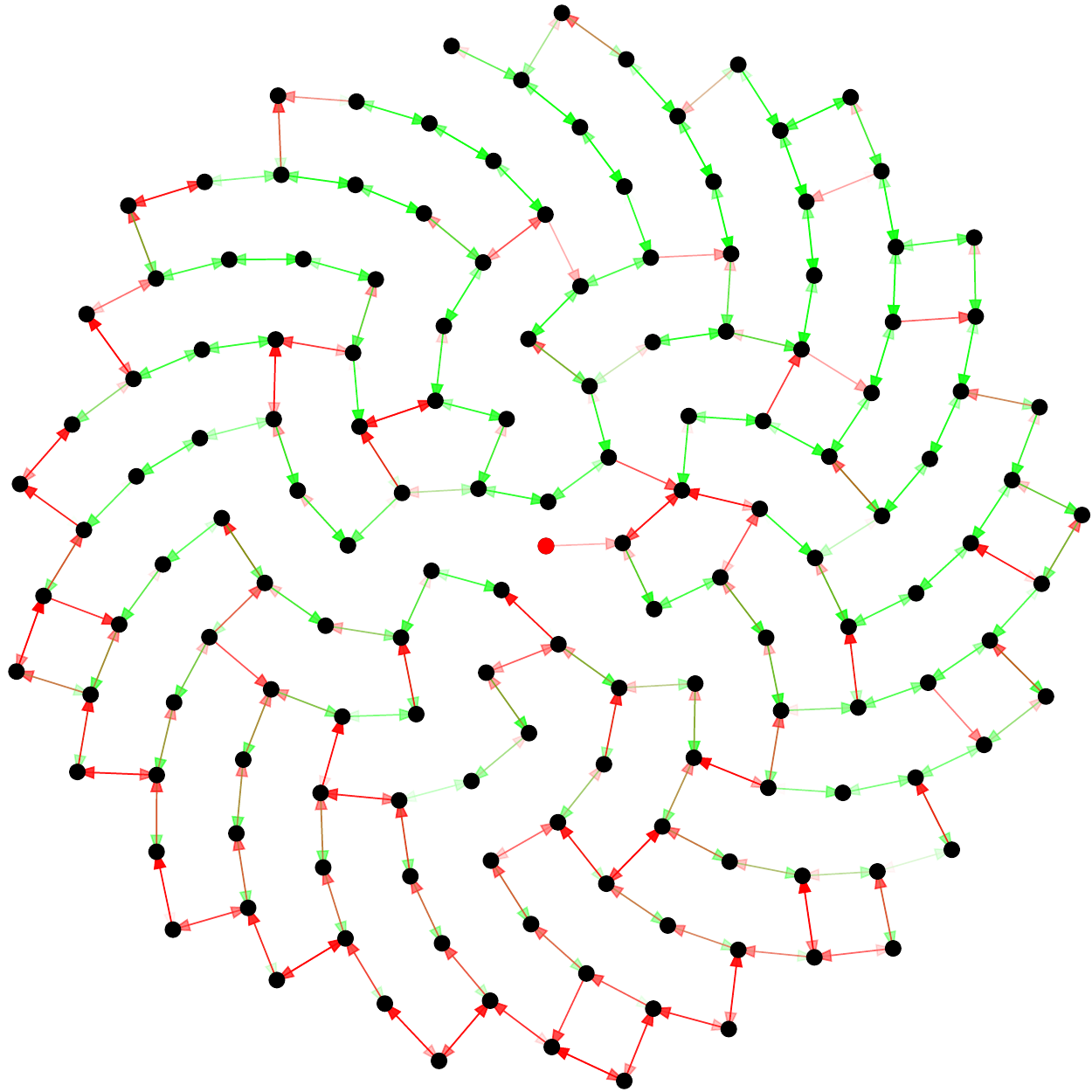} \\ 
  \end{tabular}
  \caption{Visualization of the fittest networks of each 250th generation of the most successful evolutionary run.}
  \label{fig:evolution_of_best_evolved_esn}
\end{figure}

To better understand the main difference between the best pure random echo state network and the best evolved network, their visualization is provided in Figure~\ref{fig:best_random_evolved_esn}. The random network has more than 23 thousand neural connections, in contrast with the evolved network that has only 383. The evolved network only connects neurons which are spatially close and furthermore, the network has no intersecting connections. It should be noted that the evolved network has only a single connection heading out of the input neuron and this connection has a low weight compared to the other connections in the network. The evolution of the best network is depicted in Figure~\ref{fig:evolution_of_best_evolved_esn}. We have also visualized all the other evolutionary runs and found out that the most successful networks share very similar visual features.

\subsection{Locally Connected Echo State Networks}
\label{ssec:locally_connected_esn}

A natural question emerges whether the topological features of the most successful evolved networks could be used to improve the pure random fully connected echo state networks as well. We will attempt to answer this question by restricting the pure random networks to only build local connections between the neurons. We will use the same neural substrate as in the case of evolved networks and limit the length of the connections to 0.25. Additionally, only a single connection heading out of the input neuron is allowed.

\begin{figure}
  \centering
  \includegraphics[width=0.9\linewidth]{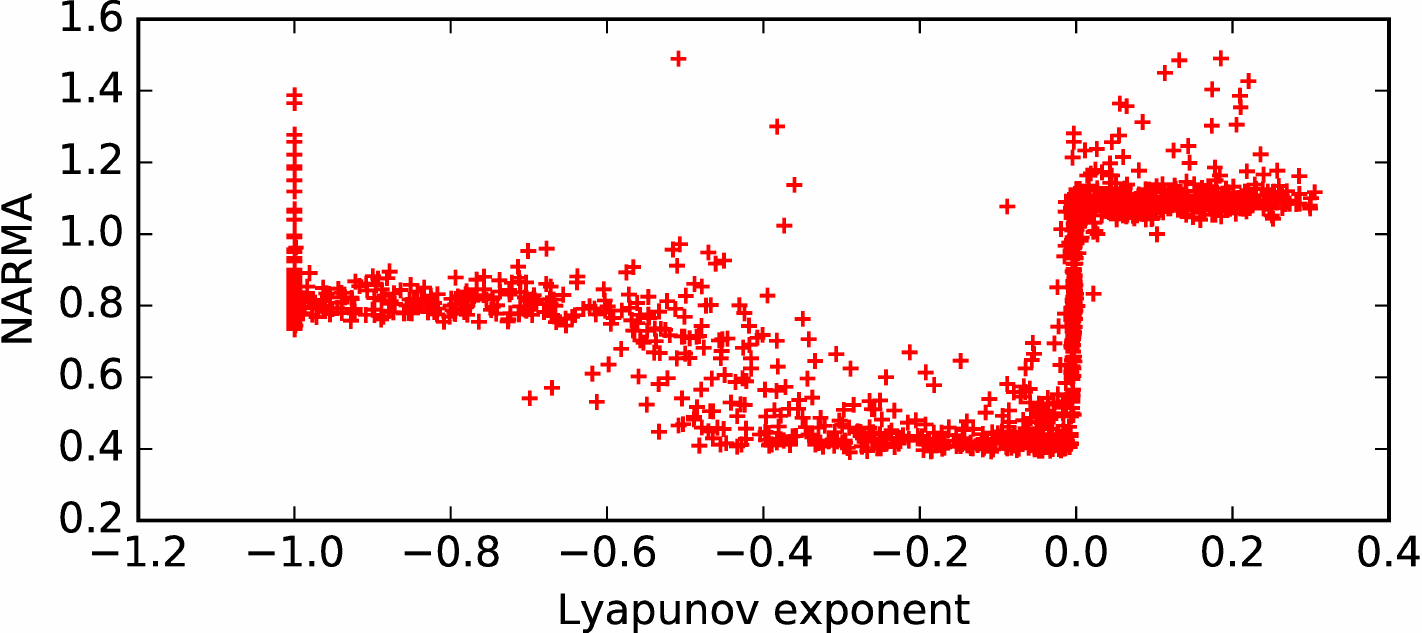}
  \caption{Plot of the NARMA error versus $\lambda$. Lower is better.}
  \label{fig:esn_local_lyap_vs_narma}
\end{figure}

According to the plot of the NARMA task in Figure~\ref{fig:esn_local_lyap_vs_narma}, the performance of the locally connected networks is again maximized on the ordered side of the edge of chaos. The other tasks manifested similar results.

\begin{figure}
  \centering
  \includegraphics[width=0.4\linewidth]{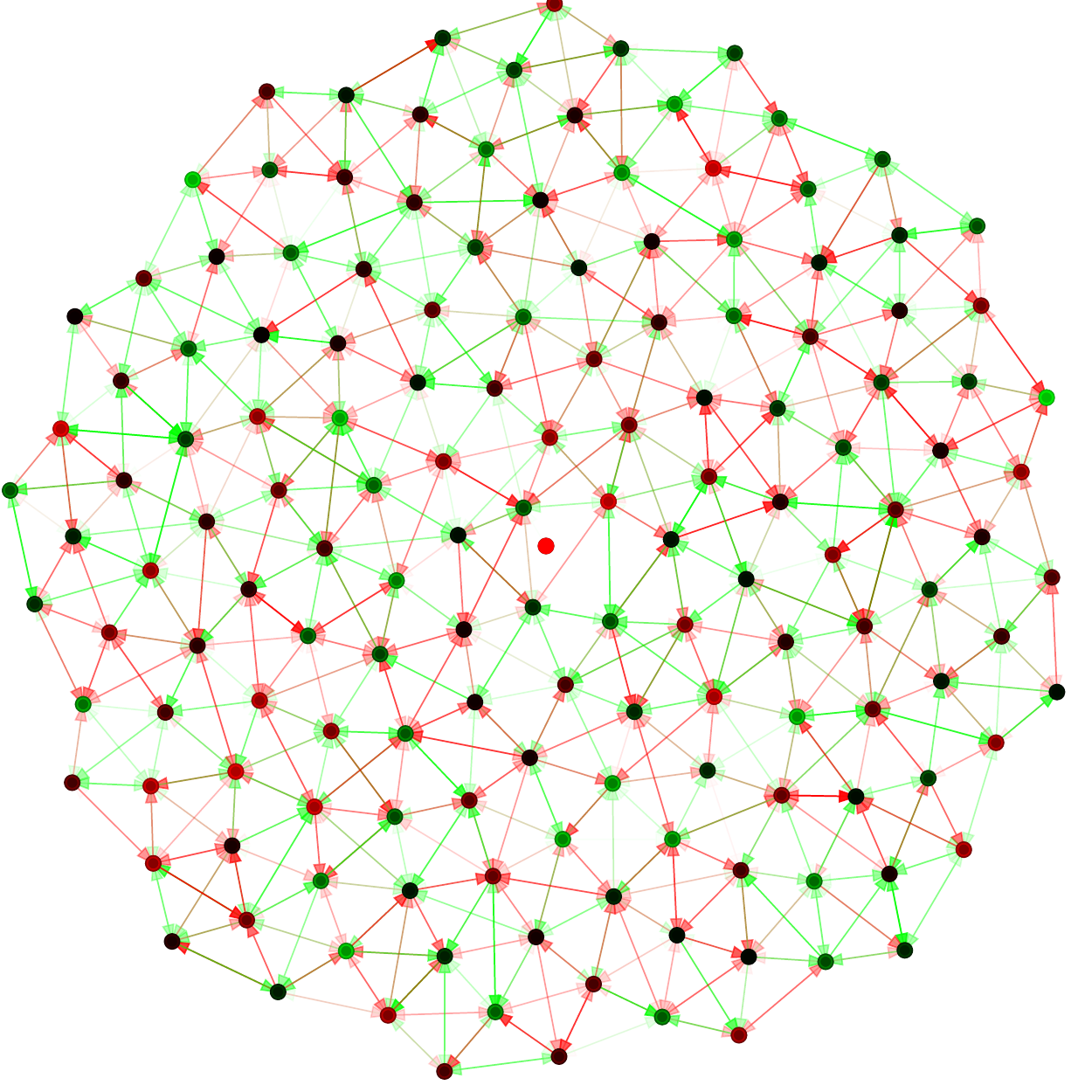}
  \caption{Illustration of the best random echo state network with local connectivity. There is a single connection heading out of the input neuron to its closest neighbor, however, it is so weak that it is barely visible.}
  \label{fig:best_local_esn}
\end{figure}

\begin{table*}
  \centering
  \setlength{\tabcolsep}{5pt}
  \begin{tabular}{lrrrr}
    \toprule
          & MMSE & NARMA & NR & Lyap. exp. \\
    \midrule
    local &  0.0104 ($\pm$ 0.0003 std) &  0.4173 ($\pm$ 0.0171 std) &  0.5325 ($\pm$ 0.0653 std) & -0.1451 ($\pm$ 0.0000 std) \\
    full  &  0.2625 ($\pm$ 0.0098 std) &  0.4150 ($\pm$ 0.0171 std) &  0.5728 ($\pm$ 0.0730 std) & -0.0280 ($\pm$ 0.0000 std) \\
    p-value &  $\boldsymbol<\mathbf{0.01}$  &  0.1318 &  $\boldsymbol<\mathbf{0.01}$ &  \\
    \bottomrule
  \end{tabular}
  \caption{The comparison of the best locally connected echo state network and the best fully connected echo state network on 50 full evaluation cycles (stabilize, train, evaluate). The p-values for the hypothesis that the performance of the locally connected network and the fully connected network differ are calculated using two-tailed one-sample t-test.}
  \label{tab:best_random_esn_vs_local}
\end{table*}

To compare locally connected and fully connected echo state networks, we again select the best candidates of both categories according to the fitness function ($2 - MMSE - NARMA$). The best locally connected network is visualized in Figure~\ref{fig:best_local_esn} and its comparison with fully connected networks is provided in Table~\ref{tab:best_random_esn_vs_local}. On the MMSE task, the locally connected network outperforms the fully connected network by one order of magnitude. On the NR task, the difference is less noticeable, yet still statistically significant. On the NARMA task, the difference is not statistically significant. The corresponding results and p-values are provided in Table~\ref{tab:best_random_esn_vs_local}. \linebreak\linebreak To summarize the results, the performance of the best locally connected network is close to the performance of the best evolved network. Locally connected networks provide a convenient alternative to neuroevolution when the available time and resources are limited.

\section{Conclusions}

Echo state networks represent a fast and powerful approach to time series analysis and prediction. However, it is difficult to choose the right set of parameters for this approach to maximize its computational performance. To simplify the parameter selection, it was stated that the performance of echo state networks is maximized when the network's dynamics is on the transition between order and chaos. We have confirmed this statement in a comprehensive set of experiments. A rigorous reason for this behaviour remains an open question.

Even though the echo state networks were designed as a model of biological brain, their fully connected topology does not appear to be biologically plausible. We have addressed this issue via evolutionary algorithms and created a network with a more ``organic'' layout. The evolved network turned out to significantly outperform the fully connected echo state networks. Furthermore, we have demonstrated that the evolution favoured the ordered side of the edge of chaos and avoided chaotic and overly ordered networks.

We have transferred the properties of the most successful evolved networks back to the original echo state networks and introduced an approach called \textit{locally connected echo state networks}. This model has also proven to significantly outperform the fully connected networks and provides a convenient alternative to neuroevolution when the computational resources are limited.

\section{Future Work}

For the comparison with other methods from the literature, both the proposed models need to be evaluated using a well known benchmark. An example of such a benchmark are the LSTM tasks defined by \citeauthor{hochreiter1997lstm} \cite{hochreiter1997lstm} on which the fully connected echo state networks have already been evaluated by \citeauthor{jaeger2012lstm} \cite{jaeger2012lstm}. Moreover, the proposed models may be evaluated on a set of real-world problems, such as speech prediction and music prediction (similarly to \citeauthor{martens2011recurrent} \cite{martens2011recurrent}).

Locally connected networks have a low number of connections with a regular structure. This opens new perspectives for an efficient \linebreak implementation using massively parallel operations. Such an implementation may allow for significantly larger networks while keeping the same computational costs.

\begin{acks}
\begin{footnotesize}

This research was supported by Charles University GA UK project number 1578717 and SVV project number 260 453.

Access to computing and storage facilities owned by parties and projects contributing to the National Grid Infrastructure MetaCentrum, provided under the programme ``Projects of Large Research, Development, and Innovations Infrastructures'' (CESNET LM2015042), is greatly appreciated.

We thank the authors of MultiNEAT (C++/Python), JIDT \cite{lizier2014jidt} (Java), and SciPy (Python) software packages for sharing their hard work under permissive open source licences.

\end{footnotesize}
\end{acks}

\bibliographystyle{ACM-Reference-Format}
\bibliography{paper} 

\end{document}